\pdfoutput=1
\documentclass[10pt]{article}
\usepackage[preprint]{tmlr}
\usepackage{amsmath,amssymb}
\usepackage{graphicx}
\usepackage{booktabs}
\usepackage{tabularx}
\usepackage{xltabular}

\usepackage{hyperref}
\usepackage{url}

\title{How Much Gradient Does Truncated Backprop Lose in Looped Transformers?}

\author{\name Aur Shalev Merin \email aur.merin@gmail.com \\
      \addr Independent researcher}

\def\month{MM}
\def\year{YYYY}
\def\openreview{\url{https://openreview.net/forum?id=XXXX}}

\begin{document}
\maketitle

\begin{abstract}
Looped transformers reuse one block of layers many times and are often trained by backpropagating through only the last $k$ loops. We measure how much of the weight gradient this throws away, exactly and for every $k$ from one backward pass, on nine looped language models, and report three findings. First, on a trained model this lost gradient is set near the output: once the loop count is well above $k$, it barely changes with the loop count (within about 0.03 from 32 to 64 loops on the three models we tested, as a pre-registered test predicted for Parcae). Where the backward signal shrinks, $\rho^{k}$, with $\rho$ the signal's fitted shrink per loop, approximates it at moderate loop counts but overstates it at long ones, because the fitted $\rho$ drifts up as the fit takes in more loops. On Ouro the signal grows, and keeping two of four loops still loses 81\% of the final output's gradient. Second, the lost gradient falls during fine-tuning, by 0.07--0.22 within 300 steps on three of four models, on held-out problems too. The fall is not just the read-out's share of the gradient; training only the layers outside the loop reproduced up to half of it, and it also falls under truncated training. So a trained model's value overstates what truncation removes later. Third, we could not show that it predicts the training cost of truncation better than rules that need no measurement: on two new models it tied Parcae's cost at the same $k$, in a test too small to separate them.
\end{abstract}

\section{Introduction}\label{sec:1}

Looped transformers reuse one block of layers many times in a row instead of stacking many different blocks. Examples are Huginn (\citealp{geiping2025}), Parcae (\citealp{prairie2026parcae}) and Ouro (\citealp{zhu2025ouro}). Running the block more times lets the model compute longer without adding parameters.

Backpropagating through every loop means storing activations for every loop. So looped models are often trained by sending the learning signal back through only the last $k$ of $R$ loops:

\begin{itemize}
\item Huginn samples the number of loops at random and backpropagates through the last 8 (\citealp[Sec.~3.3]{geiping2025});
\item Parcae uses about the last half (\citealp[Sec.~4.2]{prairie2026parcae});
\item the converted models of \citet{mcleish2025} use at most the last 8.
\end{itemize}

The choice is made by hand. Prior work has measured training cost against $k$ (\citealp[App.~I]{prairie2026parcae}; \citealp{huang2026partii}; \citealp{xiong2026truncate}), what truncation does to pretraining at fixed compute (\citealp{schwethelm2026a}), how rare, very large gradient spikes appear at large $k$ (\citealp[App.~C.1]{wang2026hrm}), and the agreement between truncated and full gradients at single settings (\citealp[App.~C.3 and C.8]{huang2026partii}). For recurrent networks unrolled over time, \citet{aicher2019} bound the bias of truncation geometrically and estimate its decay rate from data. For looped models, how much gradient truncation loses at each $k$ has not been measured.

We measure it on nine looped language models (seven published, two our own conversions), follow it into fine-tuning, and ask what it says about the cost of training with truncation. The cost results come from 600-step fine-tunes of already-trained models, where truncation costs little in absolute terms (on Parcae, keeping one of 32 loops cost about 0.04 nats and the blended settings of S5 0.005--0.016 nats, against a 0.18-nat benefit of training the loop at all). So they are a narrow proxy for choosing $k$ in pretraining. We pre-registered sixteen studies, S1--S16, listed in Appendix~\ref{app:B} (Table~\ref{tab:4}). What we find:

\begin{enumerate}
\item \textbf{On a trained model, the lost-gradient curve is set near the output, and $\rho$ drifts (Section~\ref{sec:3}).} One full backward pass per held-out sequence gives the lost gradient for every $k$. Once the loop count is well above $k$ it barely changes with the loop count (within about 0.03 at $k \leq  8$ from 32 to 64 loops; S10), as a settled forward pass predicts. Where the backward signal shrinks, $\rho^{k}$ was within about 20\% of it at the loop counts of Table~\ref{tab:1} (a range we observed, not a threshold set in advance). The fitted $\rho$ drifts up with the loop count, because a longer fit takes in more of the slowly fading far loops, related to the effect \citet[Sec.~5.4]{aicher2019} describe for recurrent networks; so $\rho^{k}$ overstates the lost gradient at longer loop counts (S10). A two-mode model gives the drift a closed form, and its pre-registered extrapolation to 96 and 128 loops held on Parcae-140m (S14), though a trend linear in log(loop count) would have passed too. The exception is Ouro, whose backward signal grows: keeping one or two of its four loops loses 92\% or 81\% of its final output's gradient, and keeping two loses 85\% under its own training objective without the entropy term (S15). These are statements about the gradient: we did not train Ouro with truncation.
\item \textbf{It falls during fine-tuning (Section~\ref{sec:4}).} After 300 full-backprop steps it was 0.07--0.22 lower per sequence on three of four models, and about as much on held-out GSM8K problems (\citealp{cobbe2021gsm8k}; S12, S13). Most of the fall remains when the gradient that gets no credit through the loop is left out (S15). A separate fine-tune that trained only the layers outside the loop, with the loop's weights fixed, reproduced about half of that loop-credited fall on two models (0.48 and 0.50) and little of it on the third (0.14; S16). That compares two runs; it does not split the full run's fall. A fine-tune through only the last 2 loops lowers it too, by 61--98\% as much (S15). With the text moving it by up to 0.12 as well, the number is a property of a checkpoint and a text, not of the model.
\item \textbf{We could not show that it predicts training cost better than rules that need no measurement (Section~\ref{sec:5}).} Within one model it is a relabelling of $k$. Across models, a pre-registered test on two new models found that a cost curve fitted on Parcae did no better than Parcae's own cost at the same $k$, the better of two such rules fixed in advance (S11), although those two models' costs differed by up to a factor of two. The test was small (six settings on two models) and could not have shown a modest advantage.
\end{enumerate}

Appendix~\ref{app:B} keeps the study-by-study record: questions, freeze commits, verdicts, amendments and exclusions. Code, pre-registrations, results and the record of when each freeze was pushed are at \url{https://github.com/Aur-Shalev-Merin/lost-gradient-looped-transformers}.

\section{Measuring the lost gradient}\label{sec:2}

\textbf{Notation.} A looped model applies the same block to a hidden state $R$ times and then reads out. Counting loops back from the output ($j = 0$ is the last loop):

\begin{itemize}
\item $\lambda_j$, the backward signal: the gradient of the loss with respect to the hidden state after loop $j$ (the adjoint);
\item $c_j$, loop $j$'s contribution to the gradient of the shared weights: $\lambda_j$ pushed back through that loop's dependence on those weights;
\item $g$, the full gradient, and $g_k$, the gradient when backprop runs through only the last $k$ loops, which drops the contributions of the loops $j \geq k$ (the layers after the loop contribute the same term to both);
\item $D_k = g - g_k$, the sum of the dropped contributions.
\end{itemize}

The lost gradient is the relative size of what truncation drops, for one held-out sequence:

\begin{equation*}\text{lost gradient}=\frac{|g-g_k|}{|g|}.\end{equation*}

It can exceed 1 when the truncated gradient points partly against the full one.

\citet{aicher2019} define the relative bias of truncated backprop as the same ratio for expected gradients, $|\mathbb{E}\,g - \mathbb{E}\,g_k| / |\mathbb{E}\,g|$. That is the quantity their convergence bound prices, and closer to what an optimiser's averaged update sees. Our tables report a different statistic: the median over held-out sequences of the per-sequence ratio, on the trained model. Section~\ref{sec:4} and Appendix~\ref{app:I} compare the two, before and during fine-tuning (S12).

\textbf{Which number.} Several protocol choices move the lost gradient by about as much as the effects we report, so we fix one and name the others where they appear. The primary statistic is the per-sequence, all-parameter ratio, as a median over held-out sequences on the trained model. It needs no batch or optimiser state, and it is what the decay rule below describes. On Parcae-140m at 32 loops and $k = 1$, the alternatives read as follows (Appendix~\ref{app:I}, Table~\ref{tab:10}):

\begin{itemize}
\item the clipped average over 64 sequences reads 0.97, against 0.80 per sequence;
\item the unclipped expected gradient of Aicher et al. loses 100\% at every $k$, because a few sequences with very large gradients dominate it;
\item after 300 fine-tuning steps the clipped averages agree with the per-sequence median within 0.05 (Section~\ref{sec:4}).
\end{itemize}

Leaving out the gradient that gets no credit through the loop raises the value by 0.01--0.04 on the trained models and by up to 0.09 after fine-tuning (Appendix~\ref{app:I}). The text moves it by up to 0.12 (Section~\ref{sec:3}), and the starred rows of Table~\ref{tab:1} use a probe over part of the loop. For what truncation removes from a training update, the per-sequence median after some fine-tuning is the closest of these. For comparing models, use one protocol throughout.

\textbf{Measuring it exactly.} One backward pass measures the lost gradient for every $k$:

\begin{enumerate}
\item Run the full backward pass once on the trained model with the weights frozen, and record the backward signal $\lambda_j$ at every loop.
\item Push each loop's signal through that loop's own weight Jacobian, one vector-Jacobian product per loop. This gives each contribution $c_j$.
\item Sum the contributions from the far end. That gives $D_k$, and so the lost gradient, at every $k$ (Appendix~\ref{app:A}).
\end{enumerate}

Equivalently, give each loop its own copy of the shared weights; standard autograd then returns every $c_j$ from one backward pass, at the price of storing a gradient per loop. Tables~\ref{tab:1} and~\ref{tab:2} use one truncated backward pass per $k$; Appendix~\ref{app:J} uses the per-loop contributions. On the same sequences the two agree within 0.001 on Parcae (fp32) and within 0.005 on Huginn (bf16). A cheaper stand-in, the summed backward signal, keeps only a running sum of the backward signal over loops, the size of a hidden state rather than of the parameters. It came within 0.01--0.03 of the loop's lost share on Parcae and Huginn, but was off by up to 0.18 on two other models (Appendix~\ref{app:H}).

\textbf{A one-number summary, $\rho$.} From the same pass, $\rho$ is the backward signal's average shrink per loop: the slope of a straight-line fit to $\log|\lambda_j|$ over all $R$ loops, as a factor, with the median taken over held-out sequences. If the signal shrinks by $\rho$ per loop and the loops push the shared weights in broadly similar directions, the dropped contributions are the tail of a geometric series, and

\begin{equation*}\text{lost gradient}\approx\rho^{k}.\end{equation*}

\begin{figure}[t]
\centering
\includegraphics[width=0.55\linewidth]{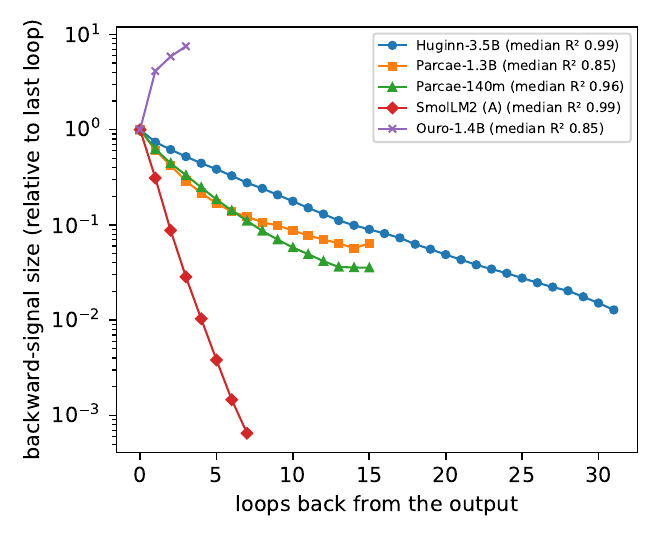}
\caption{Size of the backward signal at each loop, counting back from the output and relative to the last loop (median over sequences; log scale). The legend gives the median $R^{2}$ of a straight-line fit per sequence. Ouro's signal grows.}
\label{fig:1}
\end{figure}

With $\rho = 0.8$ and 4 loops kept, about 41\% of the gradient is lost; with $\rho = 0.35$, about 1.5\%. Appendix~\ref{app:A} gives exact forms for a finite number of loops. Geometric bounds of this kind are known (\citealp[Thm.~1]{aicher2019}; \citealp{liao2018}; \citealp[Prop.~1]{movahedi2026fprm}). \citet[Sec.~4.2]{aicher2019} also estimate the rate from data, with a log-linear fit to batch-averaged gradient norms over only the farthest 10\% of lags. We fit every loop of each sequence, a choice that matters at long loop counts (Section~\ref{sec:3}). The decay is close to geometric (Figure~\ref{fig:1}): the median $R^{2}$ of the straight-line fit is 0.99 for Huginn and SmolLM2 and 0.85--0.96 for Parcae, whose decay slows after about 8 loops (the two-mode shape of Appendix~\ref{app:J}). When the signal grows ($\rho > 1$), only the exact finite-loop forms apply.

\textbf{A dial for $\rho$.} We can slow the fade on purpose by blending each loop's output with its input, a damped iteration: $x \leftarrow (1-a)\,x + a\,F(x)$. This is the iteration \citet{geng2021} use for phantom gradients. For a loop that fully settles, it leaves the settled state, and the gradient you would get with unlimited loops, unchanged; only the per-loop shrink rate moves (Appendix~\ref{app:A}). At the finite loop counts we use, the loop has not fully settled, so the model's function changes too. The change in the trained model's loss, before any fine-tuning:

\begin{itemize}
\item Parcae-140m at 64 loops ($a = 0.3$): +0.001 on validation text, +0.02 on test text;
\item Parcae-140m at 48 loops ($a = 0.4$): +0.002 on validation text, +0.05 on test text;
\item SmolLM2 at $a = 0.5$ and 0.3: at most +0.007 on both; at $a = 0.2$, measured but never trained: +0.04.
\end{itemize}

\section{Trained models: a curve set near the output, and a drifting $\rho$}\label{sec:3}

We measured Huginn-3.5B, Parcae-1.3B, Parcae-370m, Parcae-140m, Ouro-1.4B, and two versions of SmolLM2-135M (\citealp{allal2025smollm2}) that we converted to loop some of their layers, following \citet{mcleish2025}. Five are published models from three families (Huginn, Parcae, Ouro); the two SmolLM2 conversions are our own, and one of them (B) gained nothing from looping over a never-looped control (Section~\ref{sec:7}). Two more published looped models, from \citet{huang2026partii}, appear in Appendix~\ref{app:H} and in the fine-tuning and cost tests; we call them the Part II models, after the title of that paper (``Towards Looped Models Done Right, Part II'').

\begin{table}[t]
\centering\small
\caption{Lost gradient vs the rule $\rho^{k}$ (median over held-out sequences of a single sequence's lost gradient. The rule column is the median of each sequence's $\rho_i^{k}$, so it can differ slightly from the tabulated $\rho$ raised to the power $k$).}
\resizebox{\linewidth}{!}{%
\begin{tabular}{llllll}
\toprule
model & trained with backprop through & $R$ & $\rho$ & lost gradient at $k = 1 / 2 / 4$ & rule \\
\midrule
Huginn-3.5B$^{*}$ & last 8 loops & 32 & 0.88 & 0.90 / 0.80 / 0.62 & 0.88 / 0.77 / 0.59 \\
Parcae-1.3B$^{*}$ & last half of the loops & 16 & 0.84 & 0.80 / 0.65 / 0.47 & 0.84 / 0.71 / 0.51 \\
Parcae-370m & last half of the loops & 16 & 0.81 & 0.82 / 0.67 / 0.45 & 0.81 / 0.66 / 0.43 \\
Parcae-140m & last half of the loops & 16 & 0.80 & 0.81 / 0.66 / 0.42 & 0.80 / 0.64 / 0.41 \\
SmolLM2, converted (A) & last 4 loops & 8 & 0.35 & 0.30 / 0.11 / 0.015 & 0.35 / 0.12 / 0.015 \\
SmolLM2, converted (B) & last 4 loops & 8 & 0.38 & 0.32 / 0.12 / 0.021 & 0.38 / 0.15 / 0.022 \\
Ouro-1.4B & all 4 loops\textsuperscript{\dag} & 4 & \textbf{1.88} & 0.92 / 0.81 ($k = 1$, 2) & does not fit\textsuperscript{\ddag} \\
\bottomrule
\end{tabular}}
\par\smallskip\footnotesize\raggedright
$^{*}$ Gradient measured over part of the loop (the input-injection layer and first loop layer), with a separate probe: 32 texts of 128 tokens, half WikiText (\citealp{merity2017}) and half GSM8K (\citealp{cobbe2021gsm8k}). The other rows use all parameters and 16 GSM8K validation sequences of 512 tokens, except the SmolLM2 rows, which use web text. GSM8K has no validation split; ours is the last 473 problems of its training split, held out from every fine-tune (which train on the first 7,000). Huginn samples its loop count in training; we measure at 32. Parcae's released models train with a mean of 8 loops (\citealp[Table 18]{prairie2026parcae}); we measure at 16. Table~\ref{tab:2} has the other loop counts.

\textsuperscript{\dag} The Ouro paper describes no truncation and discusses gradients ``through multiple recurrent iterations'', which implies backprop through all loops, but it does not say so explicitly. Ouro's training loss weights the losses at all 4 loops by a learned exit distribution, plus an entropy term (\citealp[Eq.~4]{zhu2025ouro}); the table measures only the final output (for the weighted loss, see ``Where the rule needs care'' below).

\textsuperscript{\ddag} Ouro's $\rho$ is a fit over 4 loops and is dominated by the last one: the signal grows about fourfold across it, and 1.25--1.37 times per loop before that (``Where the rule needs care'' below). For $\rho > 1$ the simple $\rho^{k}$ is meaningless. The exact finite-loop forms (Appendix~\ref{app:A}), computed from each sequence's $\rho_i$, give 0.92--0.99 at $k = 1$ and 0.78--0.96 at $k = 2$. The measurement lies at the lower form at $k = 1$ and between the two at $k = 2$. Every sequence has $\rho_i > 1$ (1.55 to 2.27).
\label{tab:1}
\end{table}

Two other setups appear later and give different numbers for the same models, so we name them where they are used:

\begin{itemize}
\item Section~\ref{sec:5} uses SmolLM2 (A)'s lost gradient on GSM8K text (0.36 at $k = 1$), which is what its predictions were based on; Table~\ref{tab:1} uses web text (0.30).
\item Appendix~\ref{app:E} measures Ouro with gradients averaged over clipped training batches, giving 0.996 / 0.893 at $k = 1 / 2$.
\end{itemize}

The starred rows' part-of-the-loop choice agrees with all parameters within 0.025 on Parcae-140m and SmolLM2 (A), and their probe agrees with Table~\ref{tab:1}'s method within 0.03 on Parcae-140m (Appendix~\ref{app:F}). The text moves the number more: the probe's WikiText and GSM8K halves differ by up to 0.12 at $k = 4$, and a median over 16--32 sequences has a sampling error of about 0.02--0.05.

\textbf{The lost gradient carries over across loop counts; $\rho$ does not.} Table~\ref{tab:2} shows the three models we measured at several loop counts. Once $R$ is well above $k$, the lost gradient depends on $k$ and hardly on $R$. From 32 to 64 loops it moved by at most 0.013 at $k \leq  8$ on Parcae-140m and by at most 0.011 on Parcae-370m, a pre-registered check (S10, bar 0.03). On the same 16 sequences, Huginn's moved by up to 0.034 from 32 to 48 loops, just over the same bar in a supportive check, and by up to 0.030 from 32 to 64. The fitted $\rho$ moves more: from 0.81 to 0.91 on Parcae-370m between 16 and 64 loops, from 0.77 to 0.87 on Parcae-140m between 8 and 64, and from 0.85 to 0.91 on Huginn. The rise is not universal: on the fixed-5-loop Part II model, $\rho$ falls from 0.85 at 5 loops to 0.79 at 10 (Table~\ref{tab:8}). So a lost-gradient curve measured at one loop count carries over to longer ones, and a fitted $\rho$ does not. The first is what a settled forward pass predicts: the last $k$ loops see the same states at any loop count, and the far loops added at longer loop counts contribute little even though they fade more slowly. The carry-over held to 128 loops on Parcae-140m (no change at $k \leq  8$; S14); on Parcae-370m it held to 96 loops, and at 128 its value at $k = 8$ had moved by 0.032.

\begin{table}[t]
\centering\small
\caption{The lost gradient and $\rho$ across loop counts (median over held-out sequences; all parameters for Parcae, part of the loop for Huginn).}
\begin{tabular}{lllll}
\toprule
model & $R$ & $\rho$ & lost gradient at $k = 1 / 2 / 4 / 8$ & $\rho^{k}$ at $k = 1 / 2 / 4 / 8$ \\
\midrule
Parcae-140m & 8 & 0.77 & 0.81 / 0.65 / 0.39 & 0.77 / 0.60 / 0.36 \\
Parcae-140m & 16 & 0.80 & 0.81 / 0.66 / 0.42 / 0.18 & 0.80 / 0.64 / 0.41 / 0.17 \\
Parcae-140m & 32 & 0.83 & 0.82 / 0.68 / 0.44 / 0.20 & 0.83 / 0.68 / 0.46 / 0.21 \\
Parcae-140m & 48 & 0.85 & 0.82 / 0.68 / 0.43 / 0.21 & 0.85 / 0.73 / 0.53 / 0.28 \\
Parcae-140m & 64 & 0.87 & 0.82 / 0.68 / 0.43 / 0.21 & 0.87 / 0.76 / 0.58 / 0.33 \\
Parcae-370m & 16 & 0.81 & 0.82 / 0.67 / 0.45 / 0.21 & 0.81 / 0.66 / 0.43 / 0.19 \\
Parcae-370m & 32 & 0.87 & 0.83 / 0.67 / 0.45 / 0.24 & 0.87 / 0.75 / 0.57 / 0.32 \\
Parcae-370m & 48 & 0.89 & 0.83 / 0.68 / 0.46 / 0.24 & 0.89 / 0.79 / 0.63 / 0.39 \\
Parcae-370m & 64 & 0.91 & 0.83 / 0.68 / 0.46 / 0.24 & 0.91 / 0.82 / 0.68 / 0.46 \\
Huginn-3.5B$^{*}$ & 8 & 0.85 & 0.88 / 0.75 / 0.50 & 0.85 / 0.72 / 0.53 \\
Huginn-3.5B$^{*}$ & 16 & 0.86 & 0.90 / 0.79 / 0.61 / 0.32 & 0.86 / 0.74 / 0.54 / 0.29 \\
Huginn-3.5B$^{*}$ & 32 & 0.88 & 0.90 / 0.80 / 0.62 / 0.36 & 0.88 / 0.77 / 0.59 / 0.34 \\
Huginn-3.5B$^{*}$ & 48 (16 seq.) & 0.90 & 0.92 / 0.83 / 0.65 / 0.42 & 0.90 / 0.81 / 0.66 / 0.44 \\
Huginn-3.5B$^{*}$ & 64 & 0.91 & 0.91 / 0.81 / 0.63 / 0.39 & 0.91 / 0.82 / 0.68 / 0.46 \\
\bottomrule
\end{tabular}
\par\smallskip\footnotesize\raggedright
$^{*}$ Huginn: 32 sequences at 8, 16, 32 and 64 loops; at 48 loops only the 16 sequences of S9 and S10, on which 32 loops (S9's re-run) and 64 loops give 0.91 / 0.81 / 0.64 / 0.38 and 0.92 / 0.83 / 0.63 / 0.41. Each Parcae row uses the same 16 sequences.
\label{tab:2}
\end{table}

\begin{figure}[t]
\centering
\includegraphics[width=0.55\linewidth]{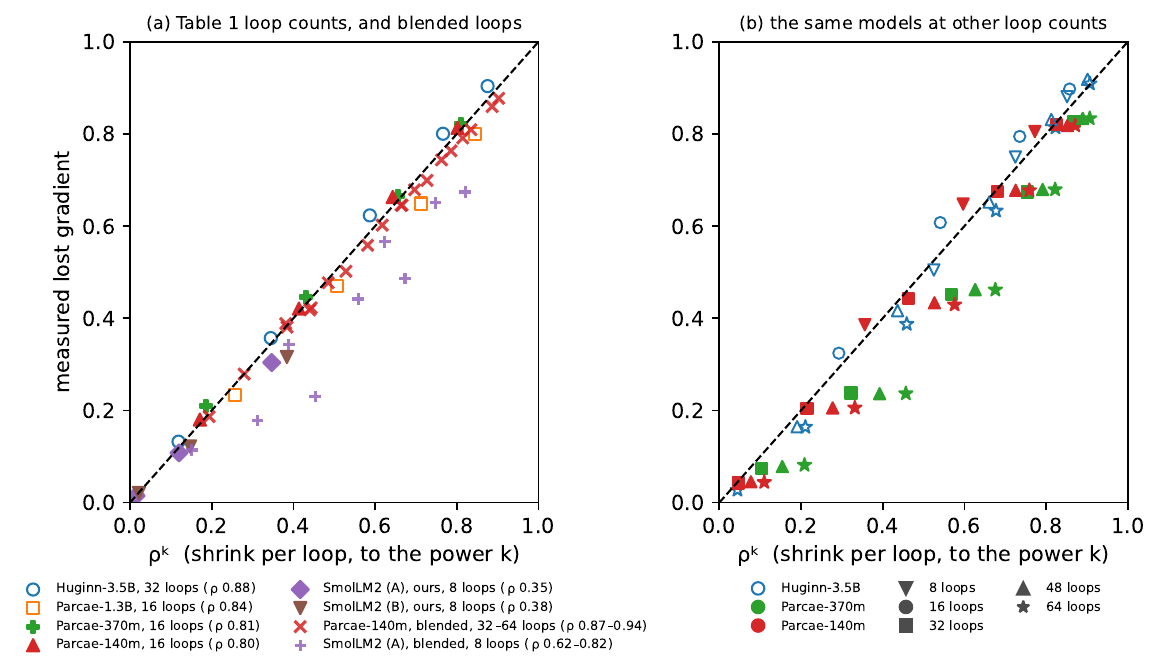}
\caption{Measured lost gradient vs $\rho^{k}$ at every measured $k$. (a) The loop counts of Table~\ref{tab:1}, for the six models whose backward signal shrinks (the SmolLM2 conversions are ours); open markers: gradient measured over part of the loop. Crosses: Parcae-140m and SmolLM2 (A) with the dial of Section~\ref{sec:2} ($a < 1$), measured with the same protocol on GSM8K validation text (Parcae-140m at $R$ 32 with $a = 0.5$ and 0.7, $R$ 48 with $a = 0.4$, $R$ 64 with $a = 0.3$; SmolLM2 (A) at $R = 8$, $a = 0.5 / 0.3 / 0.2$). (b) Parcae-140m, Parcae-370m and Huginn at their other measured loop counts, marker shape by loop count; at 48 and 64 loops (S10) the rule overstates the loss on Parcae. Ouro-1.4B is not shown ($\rho = 1.88$).}
\label{fig:2}
\end{figure}

\textbf{Figure~\ref{fig:2}} plots every measurement against $\rho^{k}$, including larger $k$ values than the tables show. In panel (a), at the loop counts of Table~\ref{tab:1}, Parcae-140m loses 0.18 at $k = 8$, against $\rho^{8} = 0.17$. $\rho$ spans 0.35 to 0.88 and the lost gradient spans 1.5\% to 90\%, yet the points sit close to the line. That span comes from three families in two clusters: the released models ($\rho$ 0.80--0.88) and our conversions (0.35--0.38), whose loops settle within a few passes (about two for conversion A; for conversion B, looping added nothing over a never-looped control; Section~\ref{sec:7}). Panel (b) shows Parcae-140m, Parcae-370m and Huginn at their other measured loop counts.

\textbf{How close is close.} For the unblended points at Table~\ref{tab:1}'s loop counts (23 points on six models), the lost gradient lies between 18\% below and 13\% above $\rho^{k}$, and within 0.07 of it (counting points with $\rho^{k} > 0.05$). At these loop counts the rule has model-specific offsets of that size: Huginn slightly above it, Parcae-1.3B and the SmolLM2 conversions slightly below. At the other loop counts we measured (51 points), the lost gradient falls up to 61\% below $\rho^{k}$ (Parcae-370m at 64 loops, $k = 16$), and already at 32 loops Parcae-370m falls 21--29\% below at $k = 4$--16. For 29 of the 68 points with per-sequence data, $\rho^{k}$ lies outside the bootstrap 95\% interval of the median (half-widths 0.01--0.19, median 0.04). We set no threshold in advance; below, ``on the line'' means inside the range at Table~\ref{tab:1}'s loop counts. These points pool Table~\ref{tab:1}'s two protocols (the starred rows use a probe over part of the loop), so the range is not a like-for-like comparison between models.

\textbf{Why $\rho^{k}$ lands close, and when it does not.} Three parts of the account were pre-registered: two here (S9, S10), and the cause of the drift in the next paragraph (S14). First, near the output successive loops' contributions point in different directions, and this turning is in the backward signal itself, not in how each loop maps it to the weights (S9, on Parcae-140m, Parcae-370m and Huginn). Second, the fitted $\rho$ follows the far loops, which fade more slowly than the loops that carry the lost gradient, so $\rho^{k}$ overshoots more as the loop count grows: at 64 loops, $\rho^{k}$ at $k = 4$ is 1.34 [1.19, 1.47] and 1.46 [1.14, 1.65] times the lost gradient on Parcae-140m and -370m (bootstrap 95\% intervals; S10). \citet[Sec.~5.4]{aicher2019} saw a related far-fade effect in recurrent networks and suggested that slowly decaying directions that do not affect the parameter gradient be ignored, leaving this to future work; the per-loop contributions do that weighting (Appendix~\ref{app:J}). The rest of the account, an exact identity that re-describes how the dropped sum shrinks and turns, and two checks that failed or were uninformative, is in Appendix~\ref{app:J}.

The drift itself has a simple cause. The backward signal near the output is the same at every loop count; adding loops only lengthens the stretch over which $\rho$ is fitted. Fitting the first 32 or 48 loops of each 64-loop profile reproduces the $\rho$ of the separate 32- and 48-loop runs (median per-sequence difference at most 0.005; the medians of $\rho$ themselves differ by up to 0.009). If the signal is a fast near-output mode plus a slow far mode, the fitted rate has a closed form (Appendix~\ref{app:J}). The two-mode curve, fitted on the 64-loop profiles, predicted $\rho$ at 96 and 128 loops on Parcae-140m before they were measured: 0.891 [0.875, 0.904] and 0.900 [0.875, 0.910], against 0.898 and 0.908 measured, with the paired rise from 64 to 128 loops above zero (+0.039 [+0.009, +0.047]; S14). That supports the extrapolation more than the specific mechanism: a trend linear in log(loop count), fitted to the 16--64-loop medians after the fact, also lands inside both intervals (0.888 and 0.903), while a straight line in the loop count does not (0.921 and 0.969). A $\rho$ fitted on a fixed window of loops next to the output would not drift by construction; we did not test whether its $\rho^{k}$ is a closer summary. The evidence for the window account itself is the window check above.

The dial of Section~\ref{sec:2} fills the gap between the two clusters (crosses in Figure~\ref{fig:2}), with mixed results. Blended Parcae-140m ($R = 32$--64, $\rho$ 0.87--0.94) sits on the line, within about 0.03 at every $k$. Blended SmolLM2 (A) ($R = 8$, $\rho$ 0.62--0.82) falls below it, by 0.10--0.22 at $a = 0.3$ and 0.2, and below both exact finite-loop forms (Appendix~\ref{app:A}), so the short loop count does not explain it; we did not measure why.

\textbf{Where the rule needs care.} On Ouro the backward signal grows: about fourfold across the last loop and by 1.2--1.4 times per loop before that, for a fitted $\rho$ of 1.88. So $\rho^{k}$ grows and the simple rule is meaningless. Ouro normalises its state between loops in its released code (\citealp[App.~D.1]{guo2026}), so the growth could have reflected the state's size rather than its sensitivity. It does not, or not only: weighting each loop's backward signal by the size of its state still gives a fitted growth of 1.66 per loop, and 1.45 on the state before normalisation (S15; median over sequences; every sequence grows on both, the smallest at 1.41 and 1.16). With only 4 loops the measurement barely constrains $\rho$, and we take one thing from Ouro: its backward signal grows, and truncation throws away most of the gradient of its final output. The same holds for the loss Ouro trains on, a weighted sum of every loop's output loss: without its entropy term, keeping 1, 2 or 3 of 4 loops loses 0.94, 0.85 or 0.73 of its gradient, slightly more than for the final output alone (0.92, 0.81, 0.71; S15). None of this measures training cost. We did not train Ouro with truncation, because it lay outside the range our cost curves were fitted on (Appendix~\ref{app:E}), and under Adam a large lost gradient can cost little (Section~\ref{sec:5}). We did not record the exit distribution, so we cannot say how much of that objective's weight falls on the last loop. Others report that Ouro's loop does not settle in the forward pass either (\citealp[Sec.~1]{huang2026partii}); those results concern the forward pass, and their link to a growing backward signal is only suggestive. Its accuracy falls past four loops (\citealp{yang2026stars}), and reusing one key-value cache slot across loop steps drops its accuracy from 78\% to 0.2\%, while Huginn is unaffected (\citealp[Table 4]{schwethelm2026b}). The warning applies to single sequences too: even at Table~\ref{tab:1}'s loop counts a few sequences have $\rho_i > 1$ (3 of 32 on Parcae-1.3B, up to 1.41; one of 16 on Parcae-140m and on Parcae-370m), and one Parcae-370m sequence loses nearly all of its gradient at every loop count we measured. The rule also fails for a loop that overshoots (Appendix~\ref{app:C}). A loop's largest stretch factor (the largest singular value of its Jacobian) is no guide: about 14 for Huginn and 30 for Parcae-1.3B, whose signals shrink, so a bound built from it says nothing useful here.

\section{The lost gradient falls during fine-tuning}\label{sec:4}

Section~\ref{sec:3} describes trained models before any further training, one sequence at a time. An optimiser sees a gradient averaged over a batch and clipped, on a model that is changing. We followed the lost gradient through fine-tuning on the four models whose truncation cost we trained (Section~\ref{sec:5}): Parcae-140m at 32 loops, SmolLM2 (A) at 8, and the two Part II models of \citet{huang2026partii} (Appendix~\ref{app:H}) at 5. Each was measured as trained and after 300 of the 600 steps of its fine-tuning recipe, with the deciding comparisons fixed in advance. Table~\ref{tab:3} gives the held-out medians; Appendix~\ref{app:I} has the details (Table~\ref{tab:10} and Table~\ref{tab:11}).

\begin{table}[t]
\centering\small
\caption{The lost gradient during fine-tuning, on held-out problems (medians over the first 64 GSM8K test problems at $k = 1 / 4$, over all parameters and with the loop-only statistic of ``Where it falls''; after 300 of 600 steps of each fine-tune; S15, S16. Table~\ref{tab:11} adds $k = 2$).}
\resizebox{\linewidth}{!}{%
\begin{tabular}{llllll}
\toprule
model & statistic & trained & full backprop & last 2 loops & no loop credit \\
\midrule
Parcae-140m & all & 0.80 / 0.41 & 0.61 / 0.22 & 0.62 / 0.21 & 0.71 / 0.34 \\
 & loop only & 0.82 / 0.42 & 0.69 / 0.25 & 0.69 / 0.24 & 0.74 / 0.35 \\
Part II, sampled loop count & all & 0.76 / 0.29 & 0.63 / 0.22 & 0.65 / 0.25 & 0.67 / 0.26 \\
 & loop only & 0.81 / 0.31 & 0.72 / 0.25 & 0.73 / 0.28 & 0.76 / 0.29 \\
Part II, fixed 5 loops & all & 0.83 / 0.39 & 0.72 / 0.30 & 0.74 / 0.37 & 0.79 / 0.38 \\
 & loop only & 0.87 / 0.40 & 0.79 / 0.33 & 0.81 / 0.40 & 0.85 / 0.40 \\
\bottomrule
\end{tabular}}
\label{tab:3}
\end{table}

\textbf{It falls, on held-out text too (S12, S13).} With full backprop, the per-sequence lost gradient fell by 0.18--0.22 on Parcae-140m, 0.07--0.16 on the Part II models and at most 0.03 on SmolLM2, at $k = 1$, 2 and 4. By the pre-set comparison, the lost share of the clipped update averaged over 64 sequences at step 300 lay 0.06--0.27 below the trained model's per-sequence median on Parcae and both Part II models. On those models, then, the frozen-model medians of Tables~\ref{tab:2} and~\ref{tab:8} overstate what truncation removes from the update at step 300 of 600, the only later stage we measured. The measured sequences had been trained on about once by then, so we repeated the measurement on GSM8K test problems, which no run trains on: as many as the split allowed, 56 batches of 8 on Parcae and 48 on the Part II models (on Parcae the repeated fine-tune used a pre-listed backup seed, Appendix~\ref{app:I}). There the fall was 90--102\% as large on average over $k$ (Parcae 0.20 / 0.22 / 0.21; the two Part II models 0.12--0.14 at $k = 1$ and 2). So the fall is not the model fitting the sequences it saw. Those problems are the same kind of text as the fine-tuning data, though, and we did not test other text.

\textbf{Where it falls (S15, S16).} The lost gradient is over all parameters, so its denominator includes gradient that gets no credit through the loop at all, in the layers after it and the read-out, and a growing share of that would lower it with no change in the loop. Leaving it out, $|g - g_k| / |g - g_0|$ with $g_0$ the gradient at $k = 0$, the loop-only held-out fall was 0.80 times the all-parameter one on Parcae and 0.72 and 0.75 times on the Part II models (averaged over $k = 1$, 2, 4; the two statistics are related by a product, not a sum, so this split is approximate). That locates the fall in the loop's part of the gradient, not its cause: a change after the loop alters the backward signal that enters it. So we ran a second fine-tune with no loop credit ($k = 0$, the baseline of Section~\ref{sec:5}), which leaves the loop's weights exactly as they were and trains only the layers after the loop and the read-out. Its loop-only lost gradient also fell, by 0.48 times the full-backprop fall on Parcae (90\% interval over sequences 0.38--0.53) and 0.50 times on the model with a sampled loop count (0.38--0.59), but by only 0.14 times on the fixed-5-loop model (0.04--0.25). So training only outside the loop reproduces about half of the fall on two models and little of it on the third. This compares two runs, not two parts of one. In the full run the outer layers adapt alongside a changing loop, and the $k = 0$ run was clipped on only 3--5\% of steps against 97--100\%. The intervals cover sequences only; a second Parcae seed gave 0.44, and the Part II models have one seed each. On Parcae the token embedding, tied to the read-out, also trained, so the loop's input changed even though its weights did not.

\textbf{It falls under truncated training too (S15).} A fine-tune with the same recipe, data order and learning rate that backpropagates through only the last 2 loops lowered the held-out lost gradient by 0.19 on Parcae (full backprop: 0.20) and by 0.10 on the model with a sampled loop count and 0.07 on the fixed-5-loop model (full backprop: 0.11 on both), averaged over $k$. On the loop-only statistic the truncated fall was 1.00, 0.80 and 0.51 times the full-backprop one (Parcae, sampled loop count, fixed 5 loops), and on the fixed-5-loop model at $k = 4$ it did not fall at all (0.400 to 0.399). The truncated gradient is smaller and was clipped on only 13--60\% of steps, so the two runs' updates differ in size.

\textbf{Averaging matters little once training is under way.} At step 300 the averaged, clipped value was within 0.05 of the per-sequence median on every model and $k$. On the trained models averaging mattered more, and not in one direction (Appendix~\ref{app:I}). Without clipping, a few sequences with very large gradients made the averaged gradient, the expected-gradient quantity of Aicher et al., lose 100\% at every $k$ on Parcae; training clipped 99--100\% of steps in all four fine-tunes, which limits those sequences' pull.

\section{What the lost gradient says about training cost}\label{sec:5}

\textbf{Setup.} We fine-tuned trained looped models on GSM8K question-and-answer text for 600 steps, with batches of 8 $\times$ 512 tokens and AdamW (\citealp{loshchilov2019}), in three kinds of run: backprop through all $R$ loops, through only the last $k$, and a baseline in which the loop gets no gradient. The training cost of truncation is the fraction of full backprop's benefit that is lost:

\begin{equation*}\text{cost}=\frac{\text{loss with $k$ loops}-\text{loss with all loops}}{\text{loss with no loop gradient}-\text{loss with all loops}}.\end{equation*}

So 0 means truncation lost nothing, and 1 means it was as bad as not training the loop. The runs being compared see the same data for the same number of steps, so this is the quality lost per training step, not at equal compute; in pretraining at fixed compute, the compute saved by truncation paid off when the loop count was fixed (\citealp{schwethelm2026a}; \citealp{xiong2026truncate}), but full backprop through fewer loops did better when the loop count could also be chosen (\citealp[Sec.~5]{xiong2026truncate}); we tested neither. All truncated and full-backprop runs of a model use one learning rate, chosen for full backprop; we did not tune it per $k$. The effects are small in absolute terms: on Parcae-140m the baseline was 0.18 nats worse than full backprop, and truncation cost 0.005 to 0.016 nats in the settings scored below. Each number averages 5 seeds (where the exclusion rule of Appendix~\ref{app:B} removed an unstable run, a pre-listed backup seed replaced it).

\textbf{Within one model, the lost gradient is a relabelling of $k$.} On a given model the lost gradient at each $k$ barely depends on the loop count (Section~\ref{sec:3}). So without an intervention, a rule using the lost gradient and a rule using $k$ alone make the same predictions, and our first study found exactly that: the cost followed the lost gradient's ordering, but not separably from $k$ (S1). The question that matters is across models.

\begin{figure}[t]
\centering
\includegraphics[width=0.55\linewidth]{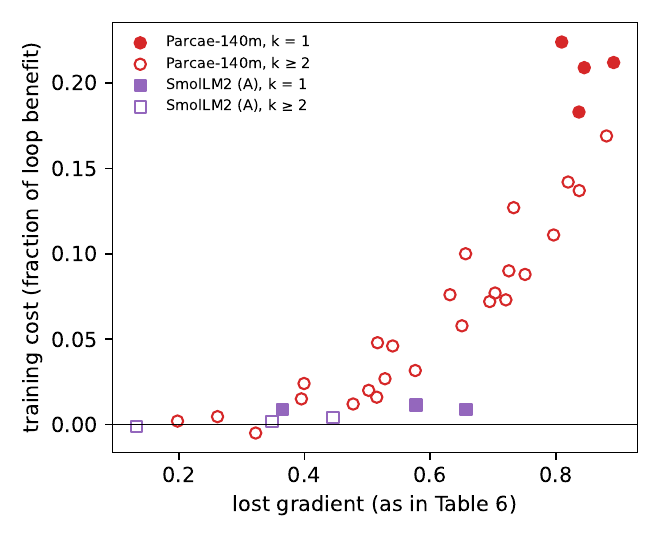}
\caption{Training cost against lost gradient for every setting we trained on Parcae-140m (29) and SmolLM2 (A) (6), lost gradient measured as in Table~\ref{tab:6} (a batch average, not the per-sequence measure the S5 and S6 curve uses). Filled markers: $k = 1$. No cutoff is implied. The two models of Appendix~\ref{app:H}, whose lost gradient was measured per sequence, are in Table~\ref{tab:9}.}
\label{fig:3}
\end{figure}

\textbf{Across models (S11).} Different models do pay different costs at the same $k$: keeping one loop costs 22\% of the loop's benefit on Parcae-140m and 1\% on SmolLM2 (A), whose loop fades much faster. A measurement of the loop could explain such differences, but these two models cannot test it: any measure of how fast the loop fades predicts near-zero cost on SmolLM2, as S6's pre-registration said in advance (Figure~\ref{fig:3}). The test needs new models on which the rules disagree before training. We ran it on the two Part II models of \citet{huang2026partii} (about 304M parameters counting embeddings, their scale S; 5 loops; Appendix~\ref{app:H}), with four predictions locked before training. Two use the frozen models' measurements and a curve fitted on Parcae before these models were measured: the lost-gradient curve of S6, and a curve in the direction of the update after Adam's rescaling (below). Two need no measurement: Parcae's own cost at the same $k$ (``$k$ alone''), and SmolLM2's. At $k = 1$ the lost-gradient curve predicted 0.14 on the model with a sampled loop count and 0.18 on the one with a fixed 5 loops, the direction-after-Adam curve 0.26 and 0.41, and the two rules without a measurement 0.22 (Parcae) and 0.01 (SmolLM2) on both.

The lost-gradient curve did no better than Parcae's own cost at the same $k$: average error 0.037 against 0.036 over the six settings ($k = 1$, 2 and 4 on both models; Table~\ref{tab:9}), so by the pre-set rule ``$k$ alone'' was as good. The direction-after-Adam curve did worse (0.051), and the two curves tied on the three settings inside both curves' fitting ranges (0.026 against 0.028). SmolLM2's near-zero costs, the other rule without a measurement, were far off (0.116), so the curve beat one of the two such rules fixed in advance and tied the other. The test had little power: a win needed the curve's average error to be half that of ``$k$ alone'' on six settings from two models, and a seed bootstrap puts the difference at $-$0.0015 to +0.0026 (Appendix~\ref{app:H}). After the fact: at the same $k$, the fixed-5-loop model, whose loop fades more slowly at its 5 loops (Table~\ref{tab:8}), cost 1.5--2.1 times as much as the other (0.29 against 0.16 at $k = 1$). No rule based on $k$ alone can produce that difference; Parcae's cost scored well by falling between the two. Both measured curves ordered the two models correctly (predicted ratios 1.3, 1.7, 2.2 for the lost-gradient curve and 1.6, 1.9, 2.0 for the direction-after-Adam curve, against 1.8, 2.1, 1.5 observed at $k = 1$, 2, 4), but the lost-gradient curve under-predicted the fixed-5-loop model's costs at $k = 1$ and 2 by about a third, and the direction-after-Adam curve over-predicted both models' costs there by 32--59\%. With one pair of models, getting the order right is weak evidence.

\begin{figure}[t]
\centering
\includegraphics[width=0.95\linewidth]{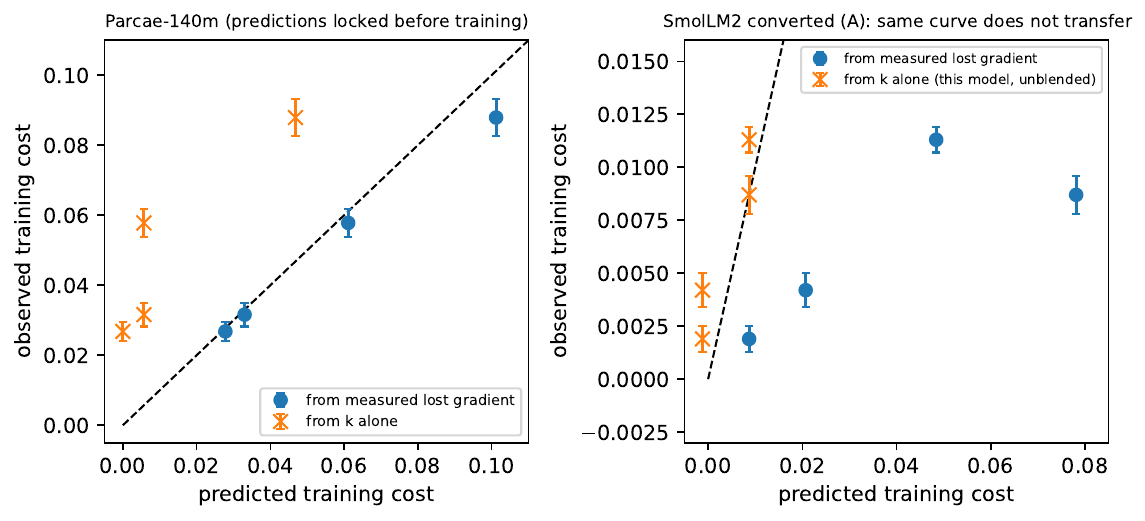}
\caption{Observed vs predicted training cost, mean $\pm$ 1 standard error over 5 seeds. Left: Parcae-140m, predictions locked before training; ``$k$ alone'' is the average cost at the same $k$ in earlier runs. Right: SmolLM2 converted (A), blended settings only (note the smaller y-axis); ``$k$ alone'' is this model's own cost without blending.}
\label{fig:4}
\end{figure}

\textbf{The earlier tests.} Before S11 we tested the lost-gradient curve within models, using the dial to change the lost gradient at a fixed $k$.

\begin{itemize}
\item \textbf{S5, Parcae-140m (Figure~\ref{fig:4}, left).} A power law fitted on 25 earlier settings, $\mathrm{cost} = 0.339\cdot(\text{lost gradient})^{3.38}$, predicted four new blended settings with an average error of 0.005, against 0.037 for ``$k$ alone''. That rival was weak, since it ignores the dial, and a rule based on $a\cdot k$ (``how far back the gradient effectively reaches''), which needs no measurement, did nearly as well (0.006 to 0.014, found after the test). The curve also works only with the measurement it was fitted on: fed batch-averaged lost gradients, its average error rises to 0.024.
\item \textbf{S6, SmolLM2 converted (A) (Figure~\ref{fig:4}, right).} The same curve, refitted with S5's results and frozen, predicted that raising the lost gradient at $k = 1$ from 0.36 to 0.65 by blending (its pre-registered measurement; Table~\ref{tab:6}'s batch average reads 0.66) would cost 5--8\%. The cost stayed at most about 1\%, and this model's own unblended cost did better (average error 0.003 against 0.032).
\end{itemize}

A reading found after the fact is consistent with S6. At $k = 1$ the dial mostly shrinks the truncated gradient instead of turning it (Appendix~\ref{app:E}), and Adam divides each weight's update by that weight's own recent gradient size, so it largely undoes a consistent shrink. At $k = 2$ and 4 on Parcae the dial also turned the direction after Adam's rescaling, and the cost rose. For SGD, a relative bias $\delta < 1$ multiplies the convergence bound by $1/(1-\delta)$ whatever its direction (\citealp[Thm.~2]{aicher2019}); our SGD runs could not check that half (Appendix~\ref{app:D}). A curve in the direction after Adam fitted both models after the fact (Appendix~\ref{app:E}), but a test within Parcae was uninformative (S8, Appendix~\ref{app:G}), and on the Part II models it did worse than ``$k$ alone'' (S11).

\textbf{What is still open.} S11 compared one pair of small models at a single loop count. Whether the lost gradient carries cost information across a wider range of models, for instance in how it orders them, needs more models than we trained. The Part II models cannot test the reading above, because blending at $k = 1$ does not act as a pure shrink on them (Appendix~\ref{app:H}). And since the lost gradient falls during fine-tuning (S12, S13), a frozen-model measurement is at best a proxy for what truncation removes during training.

\section{Related work}\label{sec:6}

\textbf{Truncated backprop.} Truncated backpropagation through time is standard for recurrent networks. \citet{aicher2019} bound its bias geometrically (Thm. 1), estimate the decay rate from late lags (Sec. 4.2), choose the truncation length adaptively, note that a few slowly decaying directions inflate the estimated rate (Sec. 5.4), and price relative bias in SGD (Thm. 2). \citet{tallec2017} remove the bias by randomising the truncation length and reweighting the backward pass, at the price of extra variance. \citet{ajalloeian2020} analyse SGD with relative gradient bias. In small recursive reasoning models truncation can even help: on ARC-AGI, leaving the first 2--3 of 8 loops without gradient raised pass@1 (\citealp[Table 3]{gao2025urm}), and on algorithmic tasks an intermediate gradient horizon generalised best (\citealp[Table 8]{shakhvalieva2026}). We measure the depth-recurrence version of this bias in looped transformers and test whether it predicts training cost.

\textbf{Fixed-point models.} For implicit and equilibrium models, a short or one-step gradient often trains well when the loop is contractive (\citealp{fung2022}; \citealp{geng2021}). \citet[Prop.~3]{liao2018} and \citet[Prop.~1]{movahedi2026fprm} bound the truncation error geometrically; Movahedi et al. state it for looped transformers but do not measure it, and their Theorem 3 (damping removes oscillation around the fixed point) underlies our Appendix~\ref{app:C}. In bilevel optimisation, \citet{shaban2019} bound the error of truncated backprop by the slowest mode, and \citet{vicol2022} note that truncation diminishes the weight on slow, low-curvature directions. Our dial is the damped iteration behind \citeauthor{geng2021}'s phantom gradients.

\textbf{Looped language models.} Huginn (\citealp{geiping2025}) and Parcae (\citealp{prairie2026parcae}) choose their backprop depth by hand; Ouro (\citealp{zhu2025ouro}) states none and apparently backpropagates through all of its loops (Table~\ref{tab:1}). \citet{mcleish2025} convert pretrained models into looped ones.

\begin{itemize}
\item Parcae \citep[App.~I]{prairie2026parcae} reports training cost against $k$.
\item HRM-Text (\citealp[App.~C.1]{wang2026hrm}) reports that backpropagating through more loops amplifies the product of loop Jacobians and brings rare, very large gradient spikes, which fits the growing backward signal we see in Ouro.
\item \citet{bansal2022} run a random number of iterations without gradient and then backpropagate through a random number of further ones, a form of truncated backprop. Together with a connection that re-injects the input, this lets recurrent networks run for many more iterations at test time without degrading.
\item \citet{schwethelm2026a} pretrain 116 models at equal total depth and fit how much one loop is worth compared with an unshared block. Backpropagating through the last half of the loops, as Parcae does, saves about 30\% of the training compute per token; at fixed compute the extra tokens lower validation loss, but each loop is worth less, which they suggest likely reflects early loops no longer receiving an accurate learning signal.
\item \citet{huang2026partii} pretrain looped models from scratch and report training cost against the backprop window (a 400M model: +15\% perplexity with one loop, within 1\% from four), the agreement between truncated and full gradients at two single settings (cosine 0.978 for the last 10 of 16 loops), and that, with the input re-injected at every loop, truncated training makes the loop settle in the forward pass (without input injection it does not; their App. C.3).
\item \citet{xiong2026truncate} fit a scaling law over nearly 600 pretraining runs that vary the loop count and the backprop window separately: at a fixed loop count truncation saves compute, but when both are chosen, full backprop through fewer loops is best, and the loss depends far more on the backprop window than on the loop count (fitted exponents 1.02 and 0.24).
\item \citet{yang2026stars} stabilise Ouro's loop past its training depth with a Jacobian penalty; \citet{guo2026} score a partial forward step $H + \alpha(F(H) - H)$ on frozen Ouro and Huginn, which at $\alpha = 0.25$ is our dial applied once.
\end{itemize}

What we add: exact measurements of the lost gradient on nine looped language models (seven published, two our own conversions), and of how fine-tuning changes it; the finding that the curve barely changes with the loop count while the fitted rate drifts (an effect known for recurrent networks, measured here in a pre-registered test); and pre-registered tests of what it predicts for training cost, which did not separate it from rules that need no measurement.

\section{Limitations and takeaways}\label{sec:7}

\textbf{Limitations.} The evidence on training cost is thin and mostly negative. No experiment here separates the lost gradient from a rule that needs no measurement: within a model, ``$k$ alone'' does as well without the dial (S1) and $a\cdot k$ nearly as well with it (S5); across models, the prediction failed under blending (S6) and, on two new models, did no better than Parcae's own cost at the same $k$ (S11). S11 used one pair of small models at 5 loops, with the token embedding frozen, so it cannot rule out a cost signal across a wider range of models. The explanation of S6's failure was found after the fact. Two choices may also favour truncation's apparent cost either way: the learning rate was tuned for full backprop and not per $k$, and S11's ``$k$ alone'' rule took Parcae's cost at the same $k$ (of 32 loops) for models run at 5 loops, rather than matching the share of loops kept.

The scope is narrow. All runs were 600-step fine-tunes, compared at the same number of steps rather than the same compute. $\rho$ needs a trained model, so it is no help when choosing $k$ for pretraining unless it is tracked during training, which we did not do; S2 and S12 show that $\rho$ and the lost gradient both move during fine-tuning. S13's held-out problems come from the fine-tuning dataset, so we did not test other text. S12 and S13 follow the full-backprop run; S15 follows one truncated run ($k = 2$) per model on three models and S16 one run with no loop credit, and other $k$ and later stages of truncated runs are unmeasured. Every model in Table~\ref{tab:1} with $\rho < 1$ was trained with truncated backprop, and the one apparently trained through all its loops (Ouro) has $\rho > 1$. Training through all loops does not by itself make the signal grow, though: the two Part II models, trained through nearly all or all of their loops, have $\rho$ of 0.80 and 0.85 (Appendix~\ref{app:H}). Ouro differs from every model whose signal shrinks in a way that may matter more: it does not re-inject the input at each loop (\citealp[Eq.~1]{zhu2025ouro}). Its prompt reaches later loops only through the loop itself, so a loop that shrank everything it passed back would also erase the prompt. This argues against $\rho$ well below 1, though not for $\rho$ as large as 1.88. Ouro also trains on a weighted loss at every loop, so its early loops get gradient from their own outputs (under such a loss a loop's own learning signal scales with its exit probability, and an early collapse of the exit gate can starve later loops (\citealp{marchenko2026})), and it normalises the state between loops (Section~\ref{sec:3}), which the Part II models do not. We tested none of these as causes; S15 shows only that the growth is not a matter of the state's size, and that the weighted loss loses as much gradient as the final output's. \citet[App.~C.3]{huang2026partii} show causally, for the forward pass, that truncated training makes a loop that re-injects its input settle (without input injection it does not, which is Ouro's case); our backward-pass pattern is consistent with that but only observed. SmolLM2's loops are also short: conversion A's unblended loop settles in about two passes, and for conversion B a version trained without ever looping did as well as the looped one.

The account of why $\rho^{k}$ lands close (Appendix~\ref{app:J}) rests on per-loop contributions of three frozen models; beyond its pre-registered parts (S9, S10, S14) it re-describes the data with an exact identity. On measurement, the starred rows of Table~\ref{tab:1} come from a different probe, and the lost gradient varies with the text by up to 0.12 (Section~\ref{sec:3}). Parcae and Huginn start the loop from a random state, drawn once per sequence with a fixed seed; we did not measure how much the lost gradient varies across draws for a fixed sequence. Last, bugs: an earlier study (not reported here) was scored and then invalidated by a precision bug that silently cut the gradient to the looped weights. Every later run checks its truncated gradient in the training precision. A second bug, in one diagnostic, was caught before scoring (Appendix~\ref{app:F}).

\textbf{Takeaways.}

\begin{itemize}
\item To check a loop for a growing backward signal, or to see what a given $k$ throws away on a given checkpoint, measure the lost gradient directly. Compare models only under one protocol: the number moves with the text (by up to 0.12), the probe and the checkpoint. One full backward pass per held-out sequence on a trained model gives it for every $k$ (for 32 sequences of Huginn-3.5B on one T4, 52 minutes at 8 loops and 2.6 hours at 64, including the truncated comparisons), and once the loop count is well above $k$ the curve measured at one loop count carries over to longer ones (to 64 loops on all three models measured, and to 128 on Parcae-140m).
\item Measure it during training too. On three of four models it fell by 0.07--0.22 within 300 full-backprop fine-tuning steps, on held-out problems from the fine-tuning dataset nearly as much as on the ones trained on, and not just through the read-out's share of the gradient; a fine-tune through only the last 2 loops lowered it too.
\item $\rho$ is a summary, not a shortcut. It needs the same full backward pass as the exact curve and saves only memory (the signal's size at each loop instead of a parameter-sized buffer per $k$). Its main use is flagging a growing signal, for single sequences as well as on average; on Ouro, keeping 2 of 4 loops still loses 81\% of the final output's gradient (85\% of its training objective's), though we did not measure what that costs Ouro in training. As an approximation, $\rho^{k}$ was within about 20\% of the lost gradient at the loop counts of Table~\ref{tab:1} (an observed range, which depends on the loop counts) and overstates it at longer loop counts.
\item In every setting we trained, small lost gradients came with small costs (Figure~\ref{fig:3}, Table~\ref{tab:9}); we set no threshold. Above that, we could not show it to be a guide to cost: a curve fitted on Parcae did no better on two other models than assuming Parcae's cost at the same $k$, and those two models' costs differed by up to a factor of two. So measure the cost with your own model and optimiser. \citet{li2026policymemory} reach the same conclusion for memory gradients in policies trained through a differentiable simulator: gradient norms measured on trained policies did not predict which cut cost more, and short training runs did. Least certain of all: which part of the gradient is lost may matter more than how much. Under Adam, a truncated gradient that was smaller but pointed nearly the same way after Adam's rescaling cost almost nothing in our runs, and one that pointed elsewhere did not. That is a lead found after the fact, and a cost curve built on it did not transfer to the Part II models.
\end{itemize}

\bibliography{main}
\bibliographystyle{tmlr}

\clearpage
\appendix

\section{The rule}\label{app:A}

Write the full gradient over the loop's parameters as $g = \sum_j c_j$, where $c_j$ is the contribution of the loop $j$ steps before the output, and keeping $k$ loops gives $g_k = \sum_{j<k} c_j$. Over all parameters, the layers after the loop add the same term to both, while the layers before it get gradient only through the loops. If the backward signal shrinks by $\rho$ per loop, $|c_j| \propto \rho^{j}$.

\begin{itemize}
\item \textbf{Aligned contributions:} lost gradient = $(\rho^{k}-\rho^{R})/(1-\rho^{R})$.
\item \textbf{Mutually orthogonal contributions:} $\sqrt{(\rho^{2k}-\rho^{2R})/(1-\rho^{2R})}$.
\item \textbf{In between:} both forms are $\approx$ $\rho^{k}$ for $\rho < 1$ and $R \gg  k$. When no two contributions point against each other, the decay rate is exactly $\rho$ and only a constant factor is uncertain: for $R \gg  k$ the lost gradient lies within a factor $\sqrt{(1+\rho)/(1-\rho)}$ of $\rho^{k}$ either way, about 3.5 at $\rho = 0.85$. Proof: write $s_j = |c_j|$, $S_k = \sum_{j\geq k} s_j$ and $Q_k = \sum_{j\geq k} s_j^2$. With no negative inner products, any set A of loops has $\sum_A s_j^2 \leq |\sum_A c_j|^2 \leq (\sum_A s_j)^2$. Applied to the dropped loops and to all loops, this gives $\sqrt{Q_k}/S_0 \leq \text{lost gradient} \leq S_k/\sqrt{Q_0}$, and with $s_j = \rho^{j}$ and $R \to \infty$ the factor above. Neither bound is attained at finite $k$; both are approached as $k$ grows. At $\rho = 0.85$ and $k = 4$ the bounds are 0.15 and 1.83. The most extreme arrangements we found (by construction and random search, not proved optimal) give 0.24--0.91, against $\rho^{4} = 0.52$: the dropped contributions all parallel and the kept ones orthogonal to everything give $S_k/\sqrt{Q_0 - Q_k + S_k^2}$ = 0.91; the dropped contributions mutually orthogonal and every kept one parallel to their sum give $\sqrt{Q_k}/(S_0 - S_k + \sqrt{Q_k})$ = 0.24.
\item \textbf{Measured-profile forms:} when the signal size $s_j$ is measured loop by loop instead of assumed geometric, the aligned and orthogonal forms become $\sum_{j\geq k} s_j / \sum_j s_j$ and $\sqrt{\sum_{j\geq k} s_j^2 / \sum_j s_j^2}$. Appendix~\ref{app:J} compares them with the measured lost gradient. In all-parameter measurements the gradient of the layers after the loop also enters the denominator. Huginn and Parcae-1.3B are measured over loop parameters only, where it is absent, and they lie above both forms too.
\item \textbf{Uneven angles:} the two forms assume every pair of contributions meets at the same angle. When the angles differ, the lost gradient can lie outside both forms with no cancellation at all. Call the coherence of a set of contributions the size of their sum divided by the sum of their sizes. Then, exactly, the lost gradient equals the aligned form on the measured sizes, $\sum_{j\geq k} |c_j| \,/\, \sum_j |c_j|$, times the coherence of the dropped loops divided by that of all loops. It lies above that form exactly when the dropped loops add up more coherently than all loops together: when they line up with each other more than the kept ones do, or when the kept and the dropped loops point in different directions. It lies below it in the reverse case. Without cancellation, the lost gradient lies between $\sqrt{\sum_{j\geq k} |c_j|^2} \,/\, \sum_j |c_j|$ and $\sum_{j\geq k} |c_j| \,/\, \sqrt{\sum_j |c_j|^2}$, a much wider range (computed with the signal sizes standing in for |$c_j$|): 0.17--0.91 for Huginn at $R = 8$, $k = 4$, and 0.16--0.80 for blended SmolLM2 (A) at $a = 0.2$. For example, kept loops all pointing one way and dropped loops pointing in unrelated directions give about 0.23 for the latter, its measured value. So a point outside both forms does not by itself show cancellation.
\item \textbf{Tail sums:} write $D_k$ for the sum of the contributions of loops $k$ and beyond, so that over the loop's parameters the lost gradient is $|D_k| / |D_0|$, and each contribution is the difference of two successive sums, $c_k = D_k - D_{k+1}$. For a loop that has settled, the tail sums are the remainder terms of a Neumann series (\citealp[App., eq.~9]{liao2018}). If the sum shrinks by a factor r per loop while turning by an angle $\theta_k$, then $|c_k| = |D_k|\sqrt{1 + r^2 - 2r\cos\theta_k}$. Turning inflates the per-loop sizes without changing what is lost. Each contribution can be computed exactly from one backward pass: loop k's recorded backward signal pushed through that loop's weight Jacobian, as in Section~\ref{sec:2}.
\item \textbf{When the rule is exact.} Write each contribution as $c_j = \rho^{j} b_j$, with the geometric shrink taken out. Then $D_k = \rho^{k} \sum_{i<R-k} \rho^{i} b_{i+k}$ and $D_0 = \sum_{i<R} \rho^{i} b_i$. Two cases make the lost gradient $\rho^{k}$ for $R \gg  k$ (in the second case its root-mean-square version), however much the contributions turn.
\begin{itemize}
\item \textbf{A fixed rotation:} if $b_{j+1} = Q b_j$ for an orthogonal Q, the first sum is $Q^{k}$ applied to the second with its last $k$ terms removed. Both sums tend to the same size as $R$ grows, so the lost gradient tends to $\rho^{k}$ in every sequence.
\item \textbf{Stationary statistics:} if the $b_j$ are random and the expected inner product of $b_i$ and $b_j$ depends only on the lag j $-$ i, the expected squared size of $\sum_{i<N} \rho^{i} b_{i+k}$ is $\sum_{i,i'<N} \rho^{i+i'} C(i'-i)$, with C the expected inner product at each lag, which does not depend on $k$. So $\mathbb{E}|D_k|^2 / \mathbb{E}|D_0|^2$ tends to $\rho^{2k}$, with corrections of order $\rho^{R-k}$. This holds for the ratio of expectations, not for each sequence or for the median over sequences that we report.
\end{itemize}
\item \textbf{$\rho > 1$:} both finite-R forms still hold. For Ouro (median $\rho = 1.88$, $R = 4$) the medians over sequences of each sequence's forms are 0.92 (aligned) and 0.99 (orthogonal) at $k = 1$ and 0.78 and 0.96 at $k = 2$.
\end{itemize}

Blending, $x \leftarrow (1-a)\,x + a\,F(x)$, maps each eigenvalue $\mu$ of the loop's Jacobian to $1 - a + a\mu$. It leaves each mode's weight, and the unlimited-loop (implicit) gradient, unchanged. Over-blending ($a > 1$) can push negative eigenvalues towards $-$1. Such a mode can then set the decay rate of the backward signal, and so $\rho$, while carrying little weight in the gradient (its weight is divided by 1 + |$\mu$|), and its alternating terms partly cancel in the first-order cost of truncation. We infer that this is what happens in Appendix~\ref{app:C}; we did not measure the per-loop contributions.

The aligned form is the ratio of two geometric sums. The orthogonal form follows because mutually orthogonal contributions satisfy $|\sum c_j|^2 = \sum |c_j|^2$. The bounds are proved above. All formulas are checked numerically on small linear loops (\texttt{code/law1\_toy\_check.py} in the code release); further results (mixtures of modes, damping) are in the supplementary material (\texttt{results\_docs/law1\_derivation\_formal.md} in the code release).

\section{Pre-registrations}\label{app:B}

All training studies on Parcae-140m used the same recipe; the SmolLM2 studies used the same recipe at $R = 8$, and S11 used it on two other models at $R = 5$, with the changes listed in Appendix~\ref{app:H}. Each training study was frozen (thresholds, analysis code and predictions committed) before its first truncated run, with one exception: S11 was frozen before its first scored run, while a three-seed pilot that included truncated runs was still running (only the pilot's seed-to-seed spread was read, to set the number of seeds). S9, S10 and S12--S16, which measure models without scoring a training run, were frozen before their first measurement.

{\small
\begin{xltabular}{\linewidth}{lXXX}
\caption{The pre-registered studies. Internal IDs are experiment numbers in our repository.}\label{tab:4}\\
\toprule
study (internal ID) & question & frozen at & outcome \\
\midrule
\endfirsthead
\multicolumn{4}{l}{\footnotesize\itshape (continued)}\\
\toprule
study (internal ID) & question & frozen at & outcome \\
\midrule
\endhead
\bottomrule
\endlastfoot
S1 (v162) & does cost follow $\rho^{k}$ ordering? & 89d8758 & consistent (ordering exact; not separable from ``$k$ alone'') \\
S2 (v163) & does raising $\rho$ by blending raise cost at matched $k$? & b9bf35c & mixed (higher at $k = 2$, 4; ties at $k = 1$, 8) \\
S3 (v164) & out-of-sample prediction, 10 settings & 1c5380c & inconclusive (closer in 9/10; missed bar due to one unstable run) \\
S4 (v165) & rerun of S3's unblended settings with an exclusion rule & 6cec94e & passed (within 0.002); a clean-up, not independent \\
S5 (v166) & Section~\ref{sec:5} test, Parcae-140m & d0a0715 & passed (average error 0.005 vs 0.037) \\
S6 (v167) & Section~\ref{sec:5} test, SmolLM2 & 880be7a (+ amendment 33fc028, made before results) & failed (``$k$ alone'' as good) \\
S7 (v170) & SGD vs Adam (Appendix~\ref{app:D}) & 29ecbb2 (+ amendment 1c67bdf, made before results) & outside planned outcomes \\
S8 (v174) & at matched agreement after Adam, do $k = 1$ and $k = 4$ cost the same? (Appendix~\ref{app:G}) & 095bd38 & uninformative (3 of 5 pairs lost their match) \\
S9 (v195) & frozen models: where does the turning come from, and do both effects of Appendix~\ref{app:J} share a source? & c1d8de0 & mixed (note below) \\
S10 (v196) & frozen models: does $\rho^{k}$ overshoot more with the loop count, and how close does the summed backward signal come? & 24a61a8 & confirmed on Parcae; the supportive Huginn check failed (note below) \\
S11 (v200) & across models: does the lost-gradient curve fitted on Parcae predict cost on two new models better than ``$k$ alone'', and better than the direction-after-Adam curve? (Appendix~\ref{app:H}) & 4849a02 & ``$k$ alone'' as good (average error 0.036 for Parcae's cost vs 0.037 for the curve); the two curves tied \\
S12 (v201) & the lost gradient of the averaged, clipped update vs the per-sequence median, on the trained model and after 300 fine-tuning steps (Appendix~\ref{app:I}) & 38b3203 (+ amendment 169d8cd, made before any Part II data) & the median overstates on Parcae and both Part II models (by 0.06--0.27); fair on SmolLM2 \\
S13 (v203) & does S12's fall during fine-tuning hold on held-out GSM8K problems? (Appendix~\ref{app:I}) & 97d945f (+ amendments 78716a1, before any data, and fc2e374, before any Parcae stage-300 data) & general: held-out fall 0.90--1.02 times the training-text fall \\
S14 (v204) & does a two-mode model of the backward signal predict the fitted $\rho$ at 96 and 128 loops on Parcae-140m? (Appendix~\ref{app:J}) & e35d92f (+ amendment f767b13, before any data) & predicted on Parcae-140m (0.898 and 0.908, inside the frozen intervals; paired rise above 0); Parcae-370m descriptive \\
S15 (v210) & is S13's fall a change in the loop, and does the lost gradient also fall along a truncated fine-tune? On Ouro, does the growth survive normalising for state size, and how much of its training objective's gradient is lost? (Sections 3 and 4, Appendix~\ref{app:I}) & 5cb861e & in the loop's part of the gradient (loop-only fall 0.72--0.80 of the all-parameter fall); falls along a $k = 2$ fine-tune (0.61--0.98 as much); Ouro grows scale-free (1.45 and 1.66); its objective loses 0.85 at $k = 2$ \\
S16 (v211) & does S15's fall need the loop's weights to change? The same fine-tune with no loop credit, which leaves them fixed (Appendix~\ref{app:I}) & 07446dd & partial: training only outside the loop gives 0.48 (Parcae), 0.50 and 0.14 (Part II) of the loop-only fall \\
\end{xltabular}}

Before S1, an earlier study was invalidated by the precision bug in Appendix~\ref{app:F}.

\textbf{S9's pre-set checks.} (A) Where the turning comes from: the backward signal itself, on all three models. (B) Whether one loop's backward map, applied repeatedly, reproduces the lost gradient at $k = 1$, 2, 4, 8 and $\rho$ within 0.02: passed on Huginn; failed on both Parcae models (Parcae-370m's lost gradient off by 0.032 at $k = 4$, Parcae-140m's $\rho$ off by 0.020). (C) Whether the turning parts carry both effects of Appendix~\ref{app:J}: uninformative, because the map is strongly non-normal. (D) An out-of-sample check on blended Parcae-140m: not scored, by design, because Parcae-140m failed (B). Computed after the fact it came within 0.03 on all 20 blended points, which is not a test. A bug in the frozen analysis code ran (D) on Huginn as well; that output was discarded.

\textbf{S10's pre-set checks.} Parcae-140m and -370m at 32, 48 and 64 loops; Huginn at 48. All three account checks passed on both Parcae models:

\begin{itemize}
\item the lost gradient at $k \leq  8$ stayed within 0.03 of its 32-loop value (largest change 0.013);
\item the fitted $\rho$ rose with the loop count;
\item $\rho^{k}$'s overshoot grew with the loop count, to 1.34 and 1.46 times the lost gradient at $k = 4$ and 64 loops (bar 1.10). The checks used point estimates; Section~\ref{sec:3} and Appendix~\ref{app:J} give bootstrap intervals.
\end{itemize}

On Huginn, a supportive check with no weight in the verdict, $\rho$ and the overshoot also rose, but its lost gradient moved by 0.034 at $k = 8$, against the 0.03 bar, so that check failed.

The summed backward signal's rate beat $\rho^{k}$ in 4 of 5 new settings, and stayed within 0.04 on average in all five; so did the summed backward signal's share. That was largely built in: both measure the lost-gradient curve itself (in the hidden state), and the rate was fitted on $k \leq  8$ and scored inside that range, while $\rho^{k}$ predicts from per-loop sizes. Fitted on $k \leq  4$ alone, after the fact, the rate ran 4--6\% low at $k = 8$ on Parcae-140m and 13--14\% low on Parcae-370m, and 10--16\% and 43--49\% low at $k = 16$; on Huginn at 48 loops it ran 5\% and 17\% high.

S11's pre-registered outcome mapping, for ``$k$ alone'' as good with Parcae's cost the best rule, said the paper would state that cost follows $k$ across models and that the lost gradient adds nothing for cost even across models. We did neither. The two Part II models' costs differed by up to a factor of two at the same $k$, which no rule based on $k$ alone can produce, and a near-tie on six settings from two models cannot show that the lost gradient adds nothing; the paper says instead that we could not show it adds anything (Section~\ref{sec:5}).

S10's pre-registered outcome mapping said Section~\ref{sec:3} would state the account as confirmed, and that the paper would recommend the summed backward signal's rate if it won the comparison. We did neither: Section~\ref{sec:3} and Appendix~\ref{app:J} separate what S10 tested from the rest of the account, and Section~\ref{sec:7} recommends measuring the lost gradient itself.

By the pre-set per-token reading rule, the turning near the output was within positions on all three models; Appendix~\ref{app:J} says what this does and does not show.

A bug in the frozen analysis code at first skipped the Huginn files; fixing it changed no verdict.

\textbf{Exclusions.} S1--S3 had no exclusion rule (S3's one unstable run is why it was inconclusive). Under the rule adopted from S4 on, S4 replaced one seed (13 unstable steps in its full-backprop run) with a pre-listed backup; S5, S6 and S11 dropped none. S13 replaced Parcae's fine-tune seed (12 unstable steps) with a pre-listed backup, by an amendment made before any Parcae stage-300 data.

\textbf{Verifiability.} The freezes are commits in our development repository, which we pushed to a private GitHub repository; it is available to reviewers on request. FREEZES.md in the code release (Section~\ref{sec:1}) lists each freeze commit with GitHub's record of when it was pushed. Only the freezes of S3, S8 and S9--S16 reached GitHub before their runs (S3's 3--9 seconds after its kernels were submitted to Kaggle, before any could start computing), so for the others the ordering of freeze and runs is our report rather than an externally timestamped record.

\textbf{$\rho$ changes with training.} In S2, every run that gave the loop gradient lowered $\rho$ by 0.03--0.05 by the end of fine-tuning, whichever $k$ it used. The lost gradient itself moves more: S12 found it 0.07--0.22 lower after 300 steps on three of four models, on held-out GSM8K problems as well (S13; Appendix~\ref{app:I}), and along a truncated fine-tune too (S15).

\section{Overshooting loops}\label{app:C}

On Parcae-140m, over-blending raises the measured $\rho$ instead of lowering it:

\begin{itemize}
\item \textbf{$a = 1.15$:} $\rho = 0.885$, and the lost gradient at $k = 4$ is 0.40, against $\rho^{4} = 0.61$.
\item \textbf{$a = 1.3$:} $\rho = 0.944$, and the model's loss also changes by 0.055.
\end{itemize}

The backward signal decays slowly in size, yet the lost gradient falls faster than $\rho^{k}$. Appendix~\ref{app:A} gives the likely reason, inferred rather than measured: over-blending makes a slow, sign-alternating mode set $\rho$, while that mode carries little of the gradient.

\section{SGD vs Adam on SmolLM2 converted (A)}\label{app:D}

\textbf{Setup.} SGD with momentum 0.9 replaced AdamW in otherwise identical runs. Learning rates for the two controls were chosen in pilots (no loop gradient: 0.3; full backprop and truncated: 0.03).

\textbf{A weak denominator.} Under SGD, the baseline with no loop gradient was only 0.026 nats worse than full backprop (0.08 under Adam), which failed the study's own pre-registered minimum gap, so costs were read through its pre-registered fallback on raw losses. The baseline also ran at a 10 times higher learning rate than the truncated runs, so costs above 1 at low learning rates partly reflect undertraining against a better-tuned baseline. Read the costs below as rough. With blending, the best truncated runs lost 58--70\% of the loop's benefit (about 0.015--0.018 nats), against about 1\% under Adam, and the best learning rate was at the top of the range tried.

\textbf{Unblended $k = 1$.} All 7 seeds diverged at learning rate 0.03. The pre-registered analysis code returned a meaningless label on the resulting NaNs, so we report the result descriptively.

\textbf{A follow-up sweep of the truncated runs} (its reading rule fixed before launch), in Table~\ref{tab:5}.

\begin{table}[!htbp]
\centering\small
\caption{SGD learning-rate sweep on SmolLM2 converted (A). Costs in units of the loop's benefit.}
\begin{tabular}{lllll}
\toprule
setting & LR 0.003 & 0.01 & 0.02 & 0.03 \\
\midrule
unblended, $k$ 1 & 3.66 & \textbf{2.18} & diverged & diverged \\
unblended, $k$ 2 & 3.85 & 3.31 & 1.56 & 1.27 \\
blended ($a = 0.3$), $k$ 1 & 4.28 & 2.14 & 1.23 & 0.70 \\
blended ($a = 0.3$), $k$ 2 & 4.14 & 1.97 & 1.07 & 0.58 \\
\bottomrule
\end{tabular}
\label{tab:5}
\end{table}

Values above 1 are worse than not training the loop. The best learning rate for the last three rows is at the edge of the grid. At a matched learning rate (0.01), blended and unblended $k = 1$ cost about the same, so these runs do not show SGD paying for a shrunk gradient.

\textbf{Parcae-140m, SGD vs Adam.} In the same study (S7), Parcae-140m at 32 loops, unblended, was fine-tuned with SGD with momentum 0.9 at learning rates chosen in a one-seed pilot (0.1 for full and truncated backprop, 0.3 for the baseline; pilot baseline gap 0.195 nats), 5 seeds. Its truncation cost was 0.26 / 0.10 / 0.01 at $k = 1 / 2 / 4$, against 0.22 / 0.10 / 0.02 under AdamW (Table~\ref{tab:6}).

\section{Size, direction, and other attempts to predict cost across models}\label{app:E}

\begin{table}[!htbp]
\centering\small
\caption{Lost gradient, its raw direction, the direction after Adam's rescaling, and training cost (frozen models, all parameters, 4 batches of 8 GSM8K validation sequences, each quantity averaged over the batches, sums in float64; ``after Adam'' divides each weight's gradient by the root mean square of that weight's gradients, with statistics from 24 clipped training batches as in D2 below; training cost as in Section~\ref{sec:5}). The lost gradient here comes from averaged batches, not Section~\ref{sec:5}'s 16 single sequences, and can read higher.}
\resizebox{\linewidth}{!}{%
\begin{tabular}{lllllllll}
\toprule
model & loops $R$ & blending $a$ & $k$ & lost gradient & size ratio $r$ & raw direction (cos) & direction after Adam & training cost \\
\midrule
SmolLM2 (A) & 8 & 1.0 & 1 & 0.36 & 0.85 & 0.935 & 0.974 & 0.009 \\
SmolLM2 (A) & 8 & 1.0 & 2 & 0.13 & 0.95 & 0.992 & 0.998 & -0.001 \\
SmolLM2 (A) & 8 & 0.5 & 1 & 0.58 & 0.50 & 0.917 & 0.975 & 0.011 \\
SmolLM2 (A) & 8 & 0.3 & 1 & 0.66 & 0.44 & 0.869 & 0.981 & 0.009 \\
SmolLM2 (A) & 8 & 0.3 & 2 & 0.44 & 0.61 & 0.962 & 0.988 & 0.004 \\
Parcae-140m & 32 & 1.0 & 1 & 0.81 & 0.46 & 0.610 & 0.793 & 0.224 \\
Parcae-140m & 32 & 1.0 & 2 & 0.65 & 0.60 & 0.775 & 0.872 & 0.100 \\
Parcae-140m & 32 & 1.0 & 4 & 0.40 & 0.79 & 0.927 & 0.940 & 0.024 \\
Parcae-140m & 48 & 0.4 & 4 & 0.75 & 0.43 & 0.678 & 0.886 & 0.088 \\
Parcae-140m & 64 & 0.3 & 8 & 0.65 & 0.52 & 0.772 & 0.918 & 0.058 \\
\bottomrule
\end{tabular}}
\par\smallskip\footnotesize\raggedright
The lost gradient mixes how much smaller the truncated gradient is (size ratio $r$) and how far its direction is off (cos); for a single batch, $\mathrm{lost}^2 = 1 + r^2 - 2r\cos$. The table averages each quantity over batches, so the identity holds only approximately for the averages.
\label{tab:6}
\end{table}

\begin{table}[!htbp]
\centering\small
\caption{Blending at $k = 1$ versus $k \geq  2$ (frozen models; lost gradient, size ratio and direction after Adam measured as in Table~\ref{tab:6}, except that the blended Parcae rows ($a = 0.5$ and 0.7) take their lost gradient and size ratio from an earlier float32 version of the same measurement, not recomputed in float64; float32 sums ran cosines 0.004--0.007 high (Appendix~\ref{app:F}), and these norm ratios may carry a bias of similar size; training cost from Section~\ref{sec:5} and S1--S3).}
\resizebox{\linewidth}{!}{%
\begin{tabular}{lllllll}
\toprule
model & $k$ & blending $a$ & lost gradient & size ratio $r$ & direction after Adam & training cost \\
\midrule
SmolLM2 (A), $R$ 8 & 1 & 1.0 / 0.5 / 0.3 & 0.36 / 0.58 / 0.66 & 0.85 / 0.50 / 0.44 & 0.974 / 0.975 / 0.981 & 0.009 / 0.011 / 0.009 \\
SmolLM2 (A), $R$ 8 & 2 & 1.0 / 0.5 / 0.3 & 0.13 / 0.35 / 0.44 & 0.95 / 0.71 / 0.61 & 0.998 / 0.988 / 0.988 & $-$0.001 / 0.002 / 0.004 \\
Parcae-140m, $R$ 32 & 1 & 1.0 / 0.5 & 0.81 / 0.89 & 0.46 / 0.28 & 0.793 / 0.790 & 0.224 / 0.212 \\
Parcae-140m, $R$ 32 & 2 & 1.0 / 0.7 / 0.5 & 0.65 / 0.73 / 0.82 & 0.60 / 0.48 / 0.37 & 0.872 / 0.859 / 0.835 & 0.100 / 0.127 / 0.142 \\
Parcae-140m, $R$ 32 & 4 & 1.0 / 0.7 / 0.5 & 0.40 / 0.54 / 0.69 & 0.79 / 0.66 / 0.52 & 0.940 / 0.926 / 0.893 & 0.024 / 0.046 / 0.072 \\
\bottomrule
\end{tabular}}
\label{tab:7}
\end{table}

\textbf{Why blending at $k = 1$ mostly shrinks the gradient.} For a settled loop, at $k = 1$ the dial changes only the size of the truncated gradient, not its direction. Near a settled state with loop Jacobian $J$, let $b$ be the last loop's direct contribution to the weight gradient. With blending $a$, the last loop contributes $a\cdot b$, while the full gradient does not depend on $a$ (Appendix~\ref{app:A}). So keeping one loop gives a smaller gradient pointing the same way: a times smaller for a loop that has fully settled, with unlimited loops. In Table~\ref{tab:7} the size ratio falls less than that, to 0.52--0.61 of its unblended value at $a = 0.3$--0.5, on both models. Our loops are finite and not exactly settled, and we did not measure the truncated and full gradients' sizes separately, so we cannot say which of the two departs from the derivation. With $k \geq  2$, blending also reweights the earlier loops against the last one, which can turn the gradient: at $k = 2$ and 4 on Parcae it turned the direction after Adam and the cost rose, while on SmolLM2 at $k = 2$ it barely did (0.998 to 0.988) and the cost rose only slightly (from $-$0.001 to 0.004, against 0.021 predicted at $a = 0.3$). This may also explain why blending did not raise Parcae's cost at $k = 1$ in an earlier study (S2). So every blended setting in S6 was one where the dial mainly shrank the gradient, and its outcome was largely built into the design. S5's settings kept 4 to 12 loops on Parcae, where blending does turn the gradient. When blending turns the gradient differs between the models; we have no simple rule for it.

The derivation holds only approximately here. SmolLM2's raw cosine at $k = 1$ does fall as blending increases (0.935, 0.917, 0.869 in Table~\ref{tab:6}), probably because with 8 loops at $a = 0.3$ the early loops are far from the settled state and the ``full'' gradient itself covers only 8 loops. Why the direction after Adam's rescaling stays flat while the raw direction falls, we cannot explain. Clipping does not remove the size difference: frozen-model gradients were 2--9 times the clip limit, and during fine-tuning the full gradient stayed above the limit on 99--100\% of steps (S12, Appendix~\ref{app:I}). The truncated gradient is smaller, though: at $k = 1$ and 2 it often falls below the limit, so the $k = 1$ and $k = 2$ runs were clipped on only 14\% and 30\% of steps in the S8 pilot on Parcae-140m (Appendix~\ref{app:G}), and clipping then shrinks the full gradient but not the truncated one. Whether the loop has settled does not separate the models either: both settle (SmolLM2 in about two passes; Parcae by $R = 32$, where truncation is still costly).

\textbf{The direction after Adam, a lead.} It ignores size by construction. It is 0.97--1.00 on SmolLM2 in every setting against 0.79--0.94 on Parcae (Tables~\ref{tab:6} and~\ref{tab:7}), and a curve in it fitted on Parcae predicted SmolLM2's costs (D2 below). The raw direction, before Adam, does not order the models: Parcae at a raw cosine of 0.93 costs 2.4\%, SmolLM2 at 0.87 costs 0.9\%. We found this measure only after trying several, could not test it on Parcae-370m (D4), and an attempt to test it within Parcae was uninformative (Appendix~\ref{app:G}). A pre-registered test on two further models (S11; Section~\ref{sec:5}, Appendix~\ref{app:H}) did not support it: its curve over-predicted their costs at $k = 1$ and 2 by 32--59\% and did worse than Parcae's own cost at the same $k$.

\textbf{Other attempts.} All were measured on frozen models, fitted on the Parcae-140m costs, and scored on SmolLM2 converted (A).

\begin{itemize}
\item \textbf{Diagnostic D1:} seven measures of how the truncated gradient compares with the full one (size, direction, sign agreement, and the progress a step makes under SGD and under Adam). None met the pre-set bar of $R^{2}$ $\geq$ 0.9 on Parcae.
\item \textbf{Diagnostic D2:} the progress lost by an Adam-like step, with statistics from 24 training batches. The pre-specified measure failed ($R^{2}$ 0.78). A secondary measure, how closely the truncated update points the same way as the full one under Adam, passed every test ($R^{2}$ 0.95), but only after the fact. We report it as an unconfirmed lead; Appendix~\ref{app:G} describes an attempt to test it within Parcae, and Appendix~\ref{app:H} a test across models.
\item \textbf{Diagnostic D3, Ouro-1.4B (no training; all parameters, gradients averaged over clipped training batches, so the values differ from Table~\ref{tab:1}'s per-sequence medians):}
\begin{itemize}
\item lost gradient 0.996 / 0.893 / 0.750 at $k = 1 / 2 / 3$ (Table~\ref{tab:1}'s protocol gives 0.92 / 0.81 / 0.71);
\item agreement of the truncated and full update directions under Adam: 0.44 / 0.66 / 0.84 (Parcae $\geq$ 0.78; SmolLM2 $\geq$ 0.97).
\item Ouro lies outside the fitted range, so by our pre-set rule we did not train it.
\end{itemize}
\item \textbf{Diagnostic D4, Parcae-370m (no training; same procedure as D2):} across 9 settings (blending $a = 1 / 0.5 / 0.3$; $k = 2 / 4 / 8$ of 32), the direction and size measures predicted costs within 0.022 of each other, short of the pre-set 0.025 needed to tell them apart. So this model cannot test the direction measure, and we did not train it.
\end{itemize}

\section{Checks and bugs}\label{app:F}

\textbf{Precision bug (an earlier study).} Under PyTorch automatic mixed precision, the loops run without gradient were computed first. They cached low-precision copies of the shared weights with no gradient connection. The later loops reused those copies, so the looped weights received almost no gradient: the truncated gradient was 87\% wrong.

The study had been scored before this was found, and it was invalidated. Every later job checks, on the GPU before training:

\begin{itemize}
\item the truncated gradient in the training precision against full precision;
\item that the written-out forward pass matches the model's own;
\item that gradient checkpointing changes nothing.
\end{itemize}

\textbf{Averaging bug (D2's first run).} Averaging raw gradients over batches let rare batches with huge gradients (norm up to 24,865 against a typical 7) dominate. Training clips any step whose gradient exceeds the limit, so the fix was to clip each batch before averaging. A sanity check flagged the problem before the run was scored.

\textbf{Long sums in float32 (an earlier version of Table~\ref{tab:6}).} Cosines summed in float32 over about $1.4\times 10^{8}$ weights on CPU ran high by 0.004--0.007, and one after-Adam cosine reached an impossible 1.017. Table~\ref{tab:6} was recomputed with all sums in float64. The after-Adam measure of D2 already used float64.

\textbf{Checking the starred rows of Table~\ref{tab:1}.} On Parcae-140m and SmolLM2 (A), from the same backward passes, we measured the lost gradient over all parameters, over the starred rows' part of the loop (input-injection layer plus first loop layer), over the whole loop, and over each loop layer alone. The part used for Huginn and Parcae-1.3B agrees with all parameters to within 0.025 at every $k$ on both models (Parcae-140m 0.82 / 0.65 / 0.40 against 0.81 / 0.66 / 0.42; SmolLM2 0.38 / 0.15 / 0.018 against 0.36 / 0.14 / 0.016, all on GSM8K validation text). A single loop layer is less reliable: across SmolLM2's eight loop layers it ranges from 0.22 to 0.41 at $k = 1$, against 0.36 over all parameters. On Ouro, the one middle layer used in earlier versions of this table reads 0.87 / 0.68 on Table~\ref{tab:1}'s sequences, below the all-parameter 0.92 / 0.81. Low precision does not matter: bfloat16, used for Huginn, moves the part-of-loop number by at most 0.015 on Parcae-140m.

The starred rows also come from a separate probe (32 texts of 128 tokens, half WikiText and half GSM8K, with its own loop code). Run on Parcae-140m in that setting, it reads 0.83 / 0.67 / 0.45 at $k = 1 / 2 / 4$, against 0.81 / 0.66 / 0.42 in Table~\ref{tab:1}, and its own $\rho$ (0.82) predicts 0.82 / 0.67 / 0.45. Given identical sequences, the two scripts give identical numbers. What does move the number is the text: the probe's WikiText and GSM8K halves differ by up to 0.12 at $k = 4$, and a median over 16--32 sequences has a sampling error of about 0.02--0.05.

\textbf{Unstable runs.} Low-precision full-backprop runs occasionally became unstable. A seed is excluded if its full-backprop run has more than 10 numerically unstable steps. The rule was fixed before it was applied.

\section{An attempted test of the direction-after-Adam lead (S8, uninformative)}\label{app:G}

\textbf{Design.} On Parcae-140m we mixed the full gradient with a $k = 1$ or a $k = 4$ truncated gradient, $g = g_\mathrm{full} + \lambda(g_k - g_\mathrm{full})$, rescaled to the full gradient's size. Five pairs of a $k = 1$ and a $k = 4$ arm were matched on the in-training agreement of Adam's update directions with a full-backprop shadow optimiser ($\bar{A}$), using one pilot seed. The pre-registered question: at matched $\bar{A}$, do the two arms cost the same (equivalence test, margin $\max(0.015,\ 0.25\times\mathrm{cost})$, adjusted for residual $\bar{A}$ differences)? 12 arms $\times$ 8 evaluation seeds, plus unscored contrasts and a pipeline check (109 runs), 600-step fine-tunes as in Section~\ref{sec:5}.

\textbf{Result: uninformative.} On the 8 scored seeds the $k = 4$ arms' agreement fell below the pilot's, so three of the five pairs failed the pre-registered drift guard ($|\text{mean}\ \Delta\bar{A}| \leq 0.01$: 0.011, 0.038 and 0.022). Of the two scored pairs, one was ``same'' and one undecided, short of the three scored pairs needed for a verdict. The pilot had matched on a single seed; the $k = 4$ arms' seed-to-seed spread near their unstable dose range was larger than assumed.

\textbf{Descriptive only.} One curve, $\mathrm{cost} = 1.33\cdot(1-\bar{A})^{0.94}$, fits all ten mixed arms of both kinds (average error 0.011). At matched $\bar{A}$, the $k = 1$ arm cost slightly more than the $k = 4$ arm in every pair (by 0.005 to 0.022, 4--31\% of the cost), inside the equivalence margin in every case. Each gap is small, but the same sign in all five pairs hints that $\bar{A}$ misses part of what sets the cost, which matters for any test built on it. An unplanned look at 81 of the 109 runs was taken before the last kernel finished; nothing was changed after it.

\section{Two more looped models: their fade, and a cost test (S11)}\label{app:H}

After the main measurements we measured two small looped models from \citet{huang2026partii} (their Part II models at the S scale, about 304M parameters), to see whether either could serve as a third model for the cost test of Section~\ref{sec:5}. Both apply a block of two transformer layers repeatedly, with the input re-injected at every loop, and both were trained through nearly all of their loops. One samples its loop count around a mean of 5 and backpropagates through the last min($R$, 10) loops; the other always runs 5 loops, with backprop through all of them. We ported their code and checked it against theirs: logits agree within 0.0001 and gradients within a relative 0.00001, and the port reproduces their WikiText perplexity (21.46 against 21.47). The protocol is Table~\ref{tab:1}'s, with each GSM8K example in the models' training format and attention kept within documents.

\begin{table}[!htbp]
\centering\small
\caption{Lost gradient in two Part II looped models (median over 16 held-out sequences, all parameters).}
\begin{tabular}{lllll}
\toprule
model & trained with backprop through & $R$ & $\rho$ & lost gradient at $k = 1 / 2 / 4$ \\
\midrule
Part II, sampled loop count & last min($R$, 10) loops & 5 & 0.80 & 0.77 / 0.60 / 0.30 \\
Part II, sampled loop count & last min($R$, 10) loops & 10 & 0.83 & 0.81 / 0.66 / 0.43 \\
Part II, fixed 5 loops & all 5 loops & 5 & 0.85 & 0.83 / 0.70 / 0.37 \\
Part II, fixed 5 loops & all 5 loops & 10 & 0.79 & 0.82 / 0.70 / 0.38 \\
\bottomrule
\end{tabular}
\label{tab:8}
\end{table}

Both backward signals shrink, and no single sequence's grows, so training through all loops does not by itself make the signal grow as it does in Ouro. At 5 loops the rule's assumption $R \gg  k$ fails, and only the finite-loop forms apply. The summed backward signal of Section~\ref{sec:2} does not carry over here: it comes within 0.03 of the loop's lost share only for the sampled-loop-count model at 10 loops, and is off by up to 0.18 elsewhere, perhaps because the input re-enters every loop. A criterion fixed before the run asked whether a model could serve for the cost test. Both qualify for a test across models: the Parcae cost curve predicts a cost at $k = 1$ between Parcae's own and SmolLM2's own. Neither can repeat Section~\ref{sec:5}'s blended contrast, because blending at $k = 1$ does not act as a pure shrink on these models.

\textbf{The cost test (S11).} It was frozen before any scored run, and the freeze was pushed to a remote server first (Appendix~\ref{app:B}). Each model was fine-tuned as in Section~\ref{sec:5} at its own 5 loops, with five arms: no loop gradient, $k = 1$, 2 and 4, and backprop through all 5 loops. Two things differ from Section~\ref{sec:5}.

\begin{itemize}
\item The learning rate was chosen on these models by the same rule: 1e-4 for full backprop, also used for every truncated arm, and 3e-4 for the no-loop-gradient baseline. Each lies inside its grid, not at an edge.
\item The token embedding was frozen in every arm. These models do not tie it to the output layer, so it gets gradient only through the loop. The baseline could not train it while the other arms could, which would have widened the baseline's gap and shrunk every cost. With it frozen everywhere, the cost compares the loop's credit alone. Full backprop then trains 205M of the 304M parameters, and the baseline 125M.
\end{itemize}

The lost gradient and the direction after Adam fed to the two curves were measured on the frozen models with the embedding frozen as well; they differ from Table~\ref{tab:8}'s all-parameter values by at most 0.006. A rule fixed in advance set the number of seeds from a three-seed pilot's spread: 5 scored seeds, with 2 pre-listed backups that were not needed. Every job first checked the truncated gradient in its training precision, that the baseline gives the loop no gradient, and that the embedding gets none. The baseline was 0.13 and 0.14 nats worse than full backprop, above the pre-set minimum of 0.03.

\begin{table}[!htbp]
\centering\small
\caption{Training cost on the two Part II models, and the four predictions locked before training (S11; cost is the mean $\pm$ standard error over 5 seeds. Predictions: the lost-gradient and direction-after-Adam curves, both fitted on Parcae, and the cost Parcae-140m and SmolLM2 (A) had at the same $k$. SmolLM2 was not trained at $k = 4$, so its rule predicts 0 there).}
\resizebox{\linewidth}{!}{%
\begin{tabular}{lllllll}
\toprule
model & $k$ & cost & lost gradient & after Adam & Parcae at same $k$ & SmolLM2 at same $k$ \\
\midrule
sampled loop count & 1 & 0.162 $\pm$ 0.002 & 0.138 & 0.258 & 0.224 & 0.009 \\
sampled loop count & 2 & 0.071 $\pm$ 0.001 & 0.058 & 0.102 & 0.100 & $-$0.001 \\
sampled loop count & 4 & 0.015 $\pm$ 0.001 & 0.005 & 0.008 & 0.024 & 0 \\
fixed 5 loops & 1 & 0.291 $\pm$ 0.002 & 0.182 & 0.412 & 0.224 & 0.009 \\
fixed 5 loops & 2 & 0.149 $\pm$ 0.002 & 0.097 & 0.197 & 0.100 & $-$0.001 \\
fixed 5 loops & 4 & 0.023 $\pm$ 0.003 & 0.011 & 0.017 & 0.024 & 0 \\
average erro$r$ &  &  & 0.037 & 0.051 & 0.036 & 0.116 \\
\bottomrule
\end{tabular}}
\label{tab:9}
\end{table}

By the pre-set rules, ``$k$ alone'' (Parcae's cost) was as good as the lost-gradient curve, and the two curves tied, both on the three settings inside both curves' fitting ranges ($k = 2$ on both models and $k = 4$ on the fixed-5-loop model; 0.026 against 0.028) and on all six (0.037 against 0.051): neither curve's average error was half the other's, the pre-set bar. On the other three settings the direction-after-Adam curve's inputs lie outside the range it was fitted on, so its predictions there are extrapolations. A seed bootstrap, after the fact, puts the lost-gradient curve's average error minus that of Parcae's cost at $-$0.0015 to +0.0026 (90\% interval, over seeds only; it does not cover variation between models). In no resample was the curve's error half that of ``$k$ alone'', which a win required.

\section{The lost gradient of the averaged update, before and during fine-tuning (S12, S13, S15, S16)}\label{app:I}

\textbf{Design} (frozen before any measurement; Appendix~\ref{app:B}).

\begin{itemize}
\item \textbf{Models:} the four whose truncation cost we trained, at their cost settings, unblended: Parcae-140m at $R = 32$, SmolLM2 converted (A) at $R = 8$, and the two Part II models at $R = 5$. The parameters are those that train in each model's study: all of them for Parcae and SmolLM2, all but the token embedding for Part II.
\item \textbf{Data:} 512 packed GSM8K training sequences of 512 tokens, as 64 batches of 8. These are the data the optimiser sees.
\item \textbf{Stages:} the trained model (stage 0), and the same model after 300 of the 600 steps of full-backprop fine-tuning with its study's recipe (stage 300).
\item \textbf{Per batch b:} $G_b$ is the gradient of the batch-mean loss with backprop through all loops, and $T_{k,b}$ the same batch and state draw with backprop through the last $k$. The lost gradient of a group of batches is $|\sum_b (G_b - T_{k,b})| / |\sum_b G_b|$. In the clipped version each $G_b$ and $T_{k,b}$ is first scaled to norm at most 1, as training's clip does.
\item \textbf{Batch sizes:} B = 8 (each batch), 64 (8 groups), 512 (all), plus B = 1 (64 single sequences, the statistic of Tables~\ref{tab:1} and~\ref{tab:2}). We report medians over groups.
\item \textbf{Decision:} the pre-set comparison was clipped B = 64 at stage 300 minus the stage-0 per-sequence median, with a bar of $\pm$0.05 at every $k$. Adam with $\beta_1 = 0.9$ at batch 8 averages about 80 sequences.
\item All measurements are in fp32, and each process checked that its full and truncated passes share their random draws.
\end{itemize}

\begin{table}[!htbp]
\centering\small
\caption{Lost gradient per sequence and of the averaged update (medians. Column heads give the stage (0 or 300 fine-tuning steps), then the statistic: ``per seq.'', the median over 64 single sequences; ``clip 64'' and ``clip 512'', the averaged update over 64 or 512 sequences with each batch of 8 clipped to norm 1 first; ``raw 512'', the same without clipping).}
\resizebox{\linewidth}{!}{%
\begin{tabular}{llllllll}
\toprule
model & $k$ & 0: per seq. & 0: clip 64 & 0: raw 512 & 300: per seq. & 300: clip 64 & 300: clip 512 \\
\midrule
Parcae-140m & 1 & 0.80 & 0.97 & 1.00 & 0.60 & 0.58 & 0.57 \\
Parcae-140m & 2 & 0.64 & 0.74 & 1.00 & 0.42 & 0.37 & 0.34 \\
Parcae-140m & 4 & 0.38 & 0.45 & 1.00 & 0.20 & 0.17 & 0.12 \\
SmolLM2 (A) & 1 & 0.36 & 0.36 & 0.36 & 0.33 & 0.31 & 0.31 \\
SmolLM2 (A) & 2 & 0.13 & 0.12 & 0.13 & 0.11 & 0.10 & 0.10 \\
SmolLM2 (A) & 4 & 0.018 & 0.013 & 0.014 & 0.012 & 0.013 & 0.010 \\
Part II, sampled loop count & 1 & 0.76 & 0.77 & 0.77 & 0.62 & 0.60 & 0.57 \\
Part II, sampled loop count & 2 & 0.60 & 0.53 & 0.59 & 0.44 & 0.42 & 0.36 \\
Part II, sampled loop count & 4 & 0.29 & 0.22 & 0.26 & 0.22 & 0.23 & 0.18 \\
Part II, fixed 5 loops & 1 & 0.83 & 0.88 & 0.82 & 0.71 & 0.69 & 0.66 \\
Part II, fixed 5 loops & 2 & 0.70 & 0.67 & 0.66 & 0.55 & 0.52 & 0.46 \\
Part II, fixed 5 loops & 4 & 0.38 & 0.29 & 0.30 & 0.29 & 0.27 & 0.22 \\
\bottomrule
\end{tabular}}
\label{tab:10}
\end{table}

\textbf{Result.} By the pre-set rule, the stage-0 per-sequence median overstates the lost share of the averaged update at stage 300 on Parcae (by 0.21--0.27), on the sampled-loop-count model (0.06--0.18) and on the fixed-5-loop model (0.11--0.18). It is fair on SmolLM2 (within 0.05). Our prediction at the freeze gave this outcome 3\%; we expected averaging, not training, to move the number, and in the other direction.

\textbf{Training.} The per-sequence lost gradient fell by 0.18--0.22 on Parcae, 0.07--0.16 on Part II and at most 0.03 on SmolLM2 between the two stages. That is more than S2's fall in $\rho$ (0.03--0.05 on held-out text over a full fine-tune, a different quantity) would suggest (Appendix~\ref{app:B}).

\textbf{Held-out check (S13).} These sequences are training data, seen about once by step 300, so S13 repeated the fine-tune and measured GSM8K test problems at both stages. On Parcae the repeated seed-2011 fine-tune failed the instability rule (12 unstable steps), and the pre-listed backup seed 2012 was used. The per-sequence fall on the test problems was 1.02, 0.90 and 0.95 times the training-sequence fall on Parcae and the two Part II models (pre-set bar 0.5: GENERAL), and seed 2013 agrees with 2012 within 0.03. The averaged, clipped update on held-out problems falls too: on Parcae, from 1.05 / 0.82 / 0.56 to 0.55 / 0.37 / 0.15. We did not test out-of-domain text.

\textbf{In the loop, and along a truncated fine-tune (S15).} S15 repeated the held-out measurement on the first 64 test problems with a second statistic, $|g - g_k| / |g - g_0|$, which leaves out the gradient that gets no credit through the loop, and added a third stage: 300 of 600 steps of the same fine-tune with backprop through the last 2 loops. Pre-set rules: the fall is in the loop if the loop-only fall is at least half the all-parameter fall on all three models, and the lost gradient falls along the truncated fine-tune if its fall is at least half the full-backprop fall on all three. Table~\ref{tab:11} gives the medians.

\begin{table}[!htbp]
\centering\small
\caption{Lost gradient over all parameters and loop only, before and after full-backprop, truncated and no-loop-credit fine-tuning (medians over 64 held-out GSM8K test problems, at $k = 1 / 2 / 4$; S15, S16).}
\begin{tabular}{llll}
\toprule
model & stage & all parameters & loop only \\
\midrule
Parcae-140m & trained & 0.80 / 0.64 / 0.41 & 0.82 / 0.66 / 0.42 \\
 & full backprop, step 300 & 0.61 / 0.43 / 0.22 & 0.69 / 0.48 / 0.25 \\
 & last 2 loops, step 300 & 0.62 / 0.43 / 0.21 & 0.69 / 0.49 / 0.24 \\
 & no loop credit, step 300 & 0.71 / 0.54 / 0.34 & 0.74 / 0.57 / 0.35 \\
Part II, fixed 5 loops & trained & 0.83 / 0.70 / 0.39 & 0.87 / 0.73 / 0.40 \\
 & full backprop, step 300 & 0.72 / 0.56 / 0.30 & 0.79 / 0.62 / 0.33 \\
 & last 2 loops, step 300 & 0.74 / 0.60 / 0.37 & 0.81 / 0.66 / 0.40 \\
 & no loop credit, step 300 & 0.79 / 0.66 / 0.38 & 0.85 / 0.71 / 0.40 \\
Part II, sampled loop count & trained & 0.76 / 0.59 / 0.29 & 0.81 / 0.63 / 0.31 \\
 & full backprop, step 300 & 0.63 / 0.46 / 0.22 & 0.72 / 0.52 / 0.25 \\
 & last 2 loops, step 300 & 0.65 / 0.47 / 0.25 & 0.73 / 0.53 / 0.28 \\
 & no loop credit, step 300 & 0.67 / 0.51 / 0.26 & 0.76 / 0.57 / 0.29 \\
\bottomrule
\end{tabular}
\label{tab:11}
\end{table}

Averaged over $k$, the loop-only fall was 0.80, 0.75 and 0.72 times the all-parameter fall, and the truncated fall 0.98, 0.61 and 0.85 times the full-backprop fall (Parcae, fixed 5 loops, sampled loop count). Both rules passed on all three models. The truncated fall is smallest on the fixed-5-loop model, whose truncated run barely moved at $k = 4$. Every run used the first pre-listed seed; Parcae's full-backprop run had 10 unstable steps, at the limit of the exclusion rule (with the same seed, S13's run had 12), and the others none. So the fine-tunes are not bit-reproducible from the seed on a GPU. The seed-2011 values agree with S13's seed 2012 (0.61 / 0.43 / 0.22 against 0.60 / 0.42 / 0.20). The truncated runs were clipped on 60\%, 46\% and 13\% of steps, against 97--100\% for full backprop.

\textbf{With the loop's weights fixed (S16).} The loop-only statistic locates the fall but cannot say what causes it, so S16 repeated the fine-tune with no loop credit ($k = 0$): same recipe, data order and learning rate, 300 of 600 steps. The loop's weights get no gradient, and the layers after the loop and the read-out train; on Parcae so does the token embedding, which is tied to the read-out, so the loop's input changes too. Pre-set rule, on r, the loop-only fall at $k = 0$ over the full-backprop one (S15's run), averaged over $k$: the fall needs the loop if r < 0.25 on all three models, training outside the loop suffices if r $\geq$ 0.75 on all three, and otherwise the result is partial. r was 0.48 on Parcae, 0.50 on the model with a sampled loop count and 0.14 on the fixed-5-loop model (all-parameter analogue 0.43, 0.62, 0.27): partial. On the fixed-5-loop model the loop-only value at $k = 4$ did not move (0.400 to 0.403). Checks set in advance passed on every model: every parameter that changed had a nonzero gradient at $k = 0$, so the loop's weights stayed exactly fixed, and the trained model's medians matched S15's exactly. All runs used the first pre-listed seed with no unstable steps; a second Parcae seed gives r = 0.44. The runs were clipped on 3--5\% of steps.

\textbf{Averaging.} Once training is under way it barely matters: at stage 300, clipped B = 64 is within 0.05 of the per-sequence median everywhere, and longer averaging lowers it a little more (clipped B = 512 is up to 0.06 below B = 64). At stage 0 it raises the value on Parcae only, where clipped B = 64 is 0.06--0.17 above the median. Unclipped, a few sequences with very large gradients dominate there and the average loses 100\% at every $k$; that unclipped average is the expected-gradient quantity of Aicher et al. On SmolLM2 and Part II the stage-0 differences are $-$0.09 to +0.05.

\textbf{Clipping.} It is on almost all the time. All four fine-tunes clipped 99--100\% of steps. Parcae's gradient norm before clipping fell from 5.6 at the first step to 1.1--1.6 from step 50 on; SmolLM2's stayed near 2.0. In the measurement batches, 91--100\% exceeded the limit at stage 300, and all of them at stage 0.

\textbf{The two Part II models.} They keep their order: the fixed-5-loop model stays above the other, by 0.07--0.10 per sequence at stage 0 and by 0.04--0.10 clipped at stage 300. So the batch-level number does not separate them more than the per-sequence one, and S11 is not re-scored with it.

\textbf{Limits.} S12 and S13 follow the full-backprop run, S15 one truncated run per model at $k = 2$, and S16 one run with no loop credit, each to step 300. The plan was amended once before any Part II data existed, because the frozen code expected a wrong parameter count (205.3M instead of 204.9M), which stopped those jobs before they measured anything.

\section{Why $\rho^{k}$ lands close: the full account}\label{app:J}

Section~\ref{sec:3} gives the pre-registered parts of this account. Here is all of it.

\textbf{The drift as a fitting-window effect (S14).} Write $s_j$ for the size of the backward signal j loops before the output. Near the output it does not depend on the loop count $R$, so the fitted $\rho$($R$) is the slope of one fixed curve, $\log s_j$, over a window that grows with $R$. Take $s_j$ as a fast and a slow mode, $A\mu_f^{\,j} + B\mu_s^{\,j}$. Its log is close to a hockey stick with slope $a_f = \log\mu_f$ up to a crossover $j^*$ and $a_s = \log\mu_s$ after it, and a least-squares line over [0, R], treating j as continuous and with $u = j^*/R \leq 1$ (for $j^* \geq R$ the fit gives $\mu_f$), then has slope

\begin{equation*}\log\rho(R) = a_s + (a_f - a_s)(3u^2 - 2u^3),\qquad u = j^*/R.\end{equation*}

With $x = j/R$ uniform on [0, 1], the slope is $a_s + (a_f - a_s)\,\mathrm{Cov}(x, -(u-x)_+)/\mathrm{Var}(x)$, and $\mathrm{Cov}(x, -(u-x)_+) = u^2/4 - u^3/6$, while $\mathrm{Var}(x) = 1/12$. The fitted rate moves from $\mu_f$ toward $\mu_s$ as $R$ grows, while the lost gradient, which lives near the output, stays put. On the 64-loop profiles the median fits are $\mu_f$ 0.75, $\mu_s$ 0.91 and $j^* \approx 17$ on Parcae-140m (0.71, 0.93 and 10 on Parcae-370m). Applied per sequence, the closed form gives medians of $\rho = 0.830 / 0.866 / 0.876$ at 32 / 48 / 64 loops, against 0.825 / 0.852 / 0.871 measured; these come from the same data and only show that the form fits. The test was S14, pre-registered: the exact least-squares slope of each sequence's fitted two-mode curve (the frozen predictor; the closed form gives nearly the same, 0.894 and 0.902) predicted $\rho$ at 96 and 128 loops on Parcae-140m as 0.891 [0.875, 0.904] and 0.900 [0.875, 0.910], measured 0.898 and 0.908, with the paired rise from 64 to 128 loops +0.039 [+0.009, +0.047] and the lost gradient unchanged. Both measured values sit about 0.008 above the predicted centre, so the two-mode fit slightly understates how far the fitted rate keeps rising. The test does not single out the two-mode model, though: a trend linear in log(loop count), fitted to the 16--64-loop medians after the fact, gives 0.888 and 0.903, also inside both intervals, while a straight line in the loop count gives 0.921 and 0.969, outside. The window check is what supports the mechanism. Parcae-370m's intervals were too wide to test (they contained its 64-loop value); it rose by +0.025 [+0.013, +0.042]. Code: \texttt{paper\_drift\_derivation.py}, \texttt{v204\_score.py}.

\textbf{The rule is not automatic.} $\rho$ comes from the backward signal through the hidden state, while the lost gradient is a sum of each loop's contribution to the weight gradient. If each contribution scales with the backward signal and none point against each other, the lost gradient shrinks at exactly the rate $\rho$, but its level can still differ from $\rho^{k}$ by a sizeable factor (the proven bounds allow a factor of about 3.5 either way at $\rho = 0.85$ when $k$ is large, and less at small $k$; Appendix~\ref{app:A}). If, in addition, every pair of contributions meets at the same angle, the lost gradient lies between the two measured-profile forms of Appendix~\ref{app:A}, computed from each sequence's measured signal sizes. It does not. On every model with a shrinking signal and every loop count for which we stored those sizes (Huginn at 8--64 loops, Parcae-1.3B at 8 and 16, Parcae-140m at 8--64, Parcae-370m at 16--64 and SmolLM2 (A)), the lost gradient is above both forms, by 0.04--0.18 at $k = 1$ and 2 (difference of medians). On Ouro, whose signal grows, it lies below both at $k = 1$.

We measured each loop's contribution directly on frozen Parcae-140m and Parcae-370m (16 and 32 loops) and Huginn (32 loops), and the backward signal at 48, 64, 96 and 128 loops (S9, S10, S14). Some of the following was pre-registered and some was read off the same data afterwards; we label which.

\emph{Pre-registered:}

\begin{itemize}
\item \textbf{The turning comes from the backward signal (S9).} Near the output, successive loops' contributions point in different directions. On all three models this turning is in the backward signal itself, not in how each loop maps it to the weights (S9, check A).
\item \textbf{$\rho^{k}$ overshoots more as the loop count grows (S10).} From 32 to 64 loops the lost gradient stayed put (Section~\ref{sec:3}) while the fitted $\rho$ rose, from 0.825 to 0.871 on Parcae-140m and from 0.868 to 0.907 on Parcae-370m. At 64 loops, $\rho^{k}$ at $k = 4$ is 1.34 [1.19, 1.47] and 1.46 [1.14, 1.65] times the lost gradient (bootstrap 95\% intervals over the 16 sequences), so the lost gradient falls 25--32\% below $\rho^{k}$ at $k = 4$ and 38--48\% below at $k = 8$. The growth from 32 loops is clear on Parcae-140m (+0.29 [+0.10, +0.35] at $k = 4$) but not resolved on Parcae-370m (+0.20 [$-$0.13, +0.36]), which already overshoots at 32 loops (1.26 [1.03, 1.47]). On Huginn, $\rho$ and the overshoot also rose, while its lost gradient moved just over the pre-set bar (Section~\ref{sec:3}). S10 tested a trend we had already seen at 16--32 loops on Parcae and 16--64 on Huginn; it confirmed the direction, not a quantitative prediction.
\item \textbf{Per token position (S10, descriptive; the reading rule was fixed in advance).} Split by token position, each position's share of the backward signal barely changes from one loop to the next (overlap 0.94--0.99), while the signal's direction at each position does (cosine 0.11--0.44 between the last two loops). This does not separate turning within a position's features from mixing across positions through attention, which could give the same pattern.
\item \textbf{Two checks did not support a single mechanism (S9).} One loop's backward map, applied repeatedly, reproduced the lost gradient and $\rho$ within 0.02 on Huginn but not on either Parcae model (off by 0.032 and 0.020). Whether the turning parts of that map carry both departures described below was uninformative: the map is strongly non-normal.
\end{itemize}

\emph{After the fact, from the same data:}

\begin{itemize}
\item \textbf{What is lost is a sum that shrinks at a steady rate.} The lost gradient is the size of $D_k$, the sum of the dropped loops' contributions. Past the first loop, that sum shrinks by a nearly constant factor per loop on Parcae-140m (0.81--0.84 at $k = 1$--16) and Huginn (0.87--0.89). On Parcae-370m it slows further back (0.82--0.83 at $k = 1$--4, 0.85--0.88 at $k = 6$--16).
\item \textbf{Near the output the sum also turns.} Each contribution is the difference of two successive sums, $c_k = D_k - D_{k+1}$, so where the sum turns, the contribution is larger than the shrinking alone would give: at the last loop 2.5 times on Parcae-140m, 2.7 times on Parcae-370m and 3.8 times on Huginn. Contribution sizes otherwise follow the backward signal (0.9--1.3 times per unit of signal, relative to the last loop), apart from Parcae's first loop, which acts on the random initial state (1.8--1.9).
\item \textbf{One identity covers both departures from the measured-profile forms.} Per-loop sizes fall fast over the first few loops before settling into the sum's rate, and the inflated near-output contributions point sideways, so they add little to the total. An exact identity (Appendix~\ref{app:A}) writes both departures in terms of one quantity, the turning of the tail sum. That is a re-description, not a tested mechanism: the S9 checks built to test a shared source failed on Parcae (B) and were uninformative (C). Why the two departures come out similar in size at these loop counts is not derived.
\item \textbf{A fit over all loops recovers the sum's rate only within a band of loop counts.} The fit is dominated by the loops where the turning has died out, so $\rho^{k}$ lands close. With few loops, the fast near-output fall pulls the fit down and $\rho^{k}$ slightly undershoots: on Parcae-140m and -370m at 16 loops, by up to about 12\% (Parcae-370m at $k = 8$). This does not hold for every model: on Parcae-1.3B at 16 loops $\rho^{k}$ lies 6--10\% above the lost gradient, on SmolLM2 at 8 loops up to 22\% above, and on Huginn at 8 loops the sign is mixed. With many loops, the fit follows the far loops, which fade more slowly than the loops that carry the lost gradient, and $\rho^{k}$ overshoots. S10 tested this last part.
\end{itemize}

\citet[Sec.~5.4]{aicher2019} saw a related far-fade effect in recurrent networks unrolled over time: a few slowly decaying directions of the backward signal pushed their estimated rate close to or above 1. They suggest that if such directions do not influence the parameter gradient they should be ignored, by weighting directions by their effect on the parameters, and leave this to future work. The per-loop contributions do that weighting, and S10's result (the lost gradient stays put while $\rho$ rises) is a measurement in looped transformers consistent with their suggestion. The ingredients of the account are also known: the tail identity (\citealp[App., eq.~9]{liao2018}), a bound set by the slowest mode (\citealp[Prop.~3.1, eq.~10]{shaban2019}), and the observation that truncation diminishes the weight on slow, low-curvature directions (\citealp[Sec.~3.3.1]{vicol2022}).

Section~\ref{sec:3} summarises the blended points (crosses in Figure~\ref{fig:2}; same measurement, same sequences). In more detail: blended Parcae-140m ($R = 32$--64, $\rho$ 0.87--0.94) sits within about 0.03 at every $k$, unlike unblended Parcae at 64 loops; we have no explanation for the difference. Blended SmolLM2 (A) ($R = 8$, $\rho$ 0.62--0.82) falls below it: by 0.10--0.22 at $a = 0.3$ and 0.2, well outside the range of Section~\ref{sec:3} (0.23 measured against $\rho^{4} = 0.45$ at $a = 0.2$), and at $a = 0.5$ by 9--12\% at $k = 1$ and 2, inside that range, but 25\% low at $k = 4$. It even falls below both exact finite-loop forms (Appendix~\ref{app:A}), by 0.02--0.10, so the short loop count does not explain it. Lying below them means that the dropped loops add up less coherently than all loops together (Appendix~\ref{app:A}), or that the contributions do not shrink the way the backward signal does; we did not measure which on this model. For this model under blending, $\rho^{k}$ overestimates the lost gradient.

\end{document}